\documentclass[10pt,journal,compsoc]{IEEEtran}
\usepackage[nocompress]{cite}

\usepackage{amsmath}
\usepackage{array}

\usepackage[caption=false,font=footnotesize,labelfont=sf,textfont=sf]{subfig}

\usepackage{url}
\usepackage{graphicx}
\usepackage{amsmath}
\usepackage{amssymb}

\usepackage{epsfig}
\usepackage{epstopdf}
\usepackage{makecell,multirow,diagbox}
\usepackage{color}
\usepackage{soul}
\usepackage{url}
\usepackage{amsmath}
\usepackage{array}
\usepackage{booktabs}

\usepackage{ragged2e}

\usepackage[utf8]{inputenc}
\usepackage{xcolor}
\usepackage{soulutf8}
\usepackage{hyperref}
\usepackage{helvet}
\usepackage{courier}
\usepackage{bm}
\usepackage{bbm}
\usepackage{wrapfig}
\usepackage{picinpar}
\usepackage{arydshln}

\usepackage[vlined,ruled,linesnumbered]{algorithm2e}
\usepackage{cite}
\usepackage{balance}
\usepackage{amssymb}

\usepackage{mathtools,xspace}

\usepackage[capitalize]{cleveref}
\crefname{section}{Sec.}{Secs.}
\Crefname{section}{Section}{Sections}
\Crefname{table}{Table}{Tables}
\crefname{table}{Tab.}{Tabs.}
\def\eg{{\it{e.g.}}}
\def\etal{{\it et al.}\xspace}

\begin{document}
\title{High-resolution Photo Enhancement in Real-time: A Laplacian Pyramid Network}

\author{
Feng~Zhang$^{\dag}$,
Haoyou~Deng$^{\dag}$,
Zhiqiang~Li,
Lida~Li,
Bin~Xu, 
Qingbo~Lu, 
Zisheng~Cao,\\
Minchen~Wei,
Changxin~Gao,
Nong~Sang,
and Xiang~Bai,~\IEEEmembership{Senior Member,~IEEE}

\IEEEcompsocitemizethanks{
\IEEEcompsocthanksitem This work was supported in part by the National Natural Science Foundation of China No.U2341227, and in part by the Hubei Provincial Natural Science Foundation of China No.2022CFA055. \textit{(Corresponding author: C.~Gao.)}
\IEEEcompsocthanksitem F.~Zhang, H.~Deng, Z.~Li, C.~Gao, and N.~Sang are with the National Key Laboratory of Multispectral Information Intelligent Processing Technology, School of Artificial Intelligence and Automation, Huazhong University of Science and Technology. E-mail: fengzhangaia@gmail.com, \{haoyoudeng, zhiqiangli, cgao, nsang\}@hust.edu.cn. 
\IEEEcompsocthanksitem L.~Li, B.~Xu, Q.~Lu and Z.~Cao are with DJI Technology Co., Ltd. E-mail: \{brown.li, mila.xu, qingbo.lu, zisheng.cao\}@dji.com.
\IEEEcompsocthanksitem M.~Wei is with Color, Imaging, and Illumination Laboratory, The Hong Kong Polytechnic University. E-mail: minchen.wei@polyu.edu.hk.
\IEEEcompsocthanksitem X.~Bai is with the School of Software Engineering, Huazhong University of Science and Technology. E-mail: xbai@hust.edu.cn.
\IEEEcompsocthanksitem $^{\dag}$ These two authors contribute equally to the work.
}
}

%

\IEEEtitleabstractindextext{%

\begin{abstract}
\justifying
Photo enhancement plays a crucial role in augmenting the visual aesthetics of a photograph. In recent years, photo enhancement methods have either focused on enhancement performance, producing powerful models that cannot be deployed on edge devices, or prioritized computational efficiency, resulting in inadequate performance for real-world applications. To this end, this paper introduces a pyramid network called LLF-LUT++, which integrates global and local operators through closed-form Laplacian pyramid decomposition and reconstruction. This approach enables fast processing of high-resolution images while also achieving excellent performance. Specifically, we utilize an image-adaptive 3D LUT that capitalizes on the global tonal characteristics of downsampled images, while incorporating two distinct weight fusion strategies to achieve coarse global image enhancement. To implement this strategy, we designed a spatial-frequency transformer weight predictor that effectively extracts the desired distinct weights by leveraging frequency features. Additionally, we apply local Laplacian filters to adaptively refine edge details in high-frequency components. After meticulously redesigning the network structure and transformer model, LLF-LUT++ not only achieves a 2.64 dB improvement in PSNR on the HDR+ dataset, but also further reduces runtime, with 4K resolution images processed in just 13 ms on a single GPU. Extensive experimental results on two benchmark datasets further show that the proposed approach performs favorably compared to state-of-the-art methods. The source code will be made publicly available at \url{https://github.com/fengzhang427/LLF-LUT}.
\end{abstract}

}

\maketitle
\IEEEdisplaynontitleabstractindextext
\IEEEpeerreviewmaketitle

\IEEEraisesectionheading{\section{Introduction}\label{sec:inntro}}

\IEEEPARstart{M}{odern} cameras, despite their advanced and sophisticated sensors, are limited in their ability to capture the same level of detail as the human eye in a given scene. In order to capture more detail, high dynamic range (HDR) imaging techniques~\cite{mantiuk2015high,banterle2017advanced} have been developed. This technology employs photo enhancement operators~\cite{oppenheim1968nonlinear,reinhard2002photographic,eilertsen2017comparative} to optimally preserve the rich details and colors of the original scene within the constrained dynamic range of display devices. However, the implementation of this operator frequently necessitates manual adjustments by experienced engineers and requires evaluation across multiple scenarios, thereby proving to be quite labor-intensive. Additionally, to further enhance the aesthetic quality of photographs, artists commonly engage in detailed manual adjustments using professional software such as Photoshop and Davinci Resolve, which also demands sophisticated photographic skills. Consequently, the automation of image quality enhancement has emerged as a focal point of research, aiming to minimize manual intervention and enhance efficiency.

In recent years, there have been notable advancements in learning-based automatic enhancement methods~\cite{yan2016automatic,gharbi2017deep, chen2018deep,park2018distort, hu2018exposure,wang2019underexposed,kosugi2020unpaired}, thanks to the rapid development of deep learning techniques~\cite{lecun2015deep}. Many of these methods focus on learning a dense pixel-to-pixel mapping between input low-quality and output high-quality image pairs. Alternatively, they predict pixel-wise transformations to map the input images. However, most previous studies involve a substantial computational burden that exhibits a linear growth pattern in tandem with the dimensions of the input image. Taking the Sony IMX586 CMOS sensor as an example, this sensor boasts a resolution of up to 48 megapixels. Processing images with such high resolution requires a massive amount of computational power, involving hundreds of billions of floating point operations (FLOPs). Moreover, storing a single image of this kind necessitates more than 20GB of memory space.

To simultaneously enhance the efficiency and performance of learning-based methods, hybrid methods~\cite{huang2019hybrid,zheng2020image,zeng2020learning,wang2021real,zhang2022clut} have emerged that combine the utilization of image priors from traditional operators with the integration of multi-level features within deep learning-based frameworks, leading to state-of-the-art performance. Among these methods, Zeng~\etal~\cite{zeng2020learning} proposes a novel image-adaptive 3-Dimensional Look-Up Table (3D LUT) based approach, which exhibits favorable characteristics such as superior image quality, efficient computational processing, and minimal memory utilization. However, as indicated by the authors, utilizing the global (spatially uniform) photo enhancement operators, such as the 3D look-up tables, may produce less satisfactory results in local areas. Additionally, this method necessitates an initial downsampling step to reduce network computations. In the case of high-resolution (4K) images, this downsampling process entails a substantial reduction factor of up to 16 times (typically downsampled to $256\times256$ resolution). Consequently, this results in a significant loss of image details and subsequent degradation in enhancement performance.

To alleviate the above problems, this work focuses on integrating global and local operators to facilitate comprehensive photo enhancement. Drawing inspiration from the reversible Laplacian pyramid decomposition~\cite{burt1987Laplacian} and the classical local photo enhancement operators, the local Laplacian filter~\cite{paris2011local,aubry2014fast}, we propose a real-time end-to-end framework, denoted as LLF-LUT++, for high-resolution photo enhancement performing global tone manipulation while preserving local edge details. It processes images at a 4K resolution on a single GPU in just 13 ms.
Specifically, we develop a lightweight spatial-frequency transformer weight predictor to predict the global-level weight points and pixel-level content-dependent weight maps. These are used to refine the high-resolution (HR) input image and its low-resolution (LR) counterpart, respectively. The combined weight fusion strategy can produce visually pleasing results with minimal computational cost. To preserve local edge details and reconstruct the image from the Laplacian pyramid faithfully, we propose an image-adaptive learnable local Laplacian filter (LLF) to refine the high-frequency components while minimizing the use of computationally expensive convolution in the high-resolution components for efficiency. Consequently, we progressively construct a compact network to learn the parameter value maps at each level of the Laplacian pyramid and apply them to the remapping function of the local Laplacian filter. Moreover, a fast local Laplacian filter~\cite{aubry2014fast} is employed to replace the conventional local Laplacian filter~\cite{paris2011local} for computational efficiency. With these effective designs, this framework achieves real-time processing for 4K images. Extensive experimental results on two benchmark datasets demonstrate that the proposed method performs favorably against state-of-the-art methods.

In conclusion, the highlights of this work can be summarized into four points:

(1) We present an end-to-end framework for real-time high-resolution photo enhancement. This network simultaneously performs global tone manipulation and local edge detail preservation within a unified model.

(2) We integrate two distinct weight fusion strategies to achieve coarse global tone manipulation. By developing a spatial-frequency transformer weight predictor, we enhance images with minimal computational overhead.

(3) We propose an image-adaptive, learnable local Laplacian filter for efficient preservation of local edge details, which demonstrates remarkable effectiveness when integrated with an image-adaptive 3D LUT.

(4) We conduct extensive experiments on two publically available benchmark datasets. Both qualitative and quantitative results demonstrate that the proposed method performs favorably against state-of-the-art methods.

This work is an extension of our earlier conference version that has appeared in NeurIPS 2023~\cite{zhang2024lookup}. In comparison to the conference version, we have introduced a significant amount of new materials.
1) We introduce a novel LUT fusion approach for coarse global tone adjustments. Unlike LLF-LUT, which employs a weight-map-based 3D-LUT fusion, our method integrates two distinct fusion strategies, improving both efficiency and performance.
2) We design a new weight predictor to enhance the original framework. LLF-LUT directly adopts an existing transformer architecture for weight prediction, which is suboptimal for image enhancement. To address this, we design a specialized transformer capable of generating distinct LUT weights, better aligning with the fusion strategy requirements.
3) We achieve a high-quality enhancement for high-resolution images in real time. Despite reducing the parameter count by 14K, our approach improves PSNR by 2.64 dB on the HDR+ dataset compared to LLF-LUT. Additionally, by redesigning the model architecture, we reduce the processing time for 4K images from 20.51 ms to 13.50 ms.
4) Extensive experiments, design analyses, and ablation studies were conducted to demonstrate the effectiveness of our method over existing state-of-the-art techniques. Furthermore, we provide a more thorough literature review on photo enhancement, discussing the strengths and limitations of existing methods.

\section{Related Work}
\label{related}
Our work is an attempt for photo enhancement by combining the 3D look-up table with the local Laplacian filter, which is rarely touched in the previous works.
Traditional methods rely on example-based references and manually applied operators to transfer color and style, limiting their ability to process images automatically. With the advent of deep learning, learning-based approaches have enabled automated photo enhancement, achieving high performance yet incurring significant computational costs.
To improve efficiency, hybrid methods that leverage both neural networks and traditional operators have gained attention. Typically, 3D-LUT based approaches have demonstrated superior performance while requiring only shallow networks and minimal computational resources.
In what follows, we provide a review of example-based methods, learning-based methods, and 3D-LUT-based approaches.

\vspace{0.5em}
\noindent
\textbf{Example-based photo enhancement.}
Example-based photo enhancement is aimed at transferring the color and style of the reference image to the target image. Traditionally, the features that characterize the example image and the operators that perform the style transformation are obtained manually. These traditional operators can be classified according to their processing as global or local. Global operators~\cite{ferwerda1996model,reinhard2001color,reinhard2002photographic,drago2003adaptive,reinhard2005dynamic,kuang2007icam06,pitie2005n,pitie2007linear,tai2007soft} map each pixel according to its global characteristics, irrespective of its spatial localization. This approach entails calculating a single matching luminance value for the entire image. As a result, the processing time is considerably reduced, but the resulting image may exhibit fewer details. In contrast, local operators~\cite{debevec2002tone,durand2002fast,fattal2002gradient,li2005compressing,paris2011local,pouli2011progressive} consider the spatial localization of each pixel within the image and process them accordingly. In essence, this method calculates the luminance adaptation for each pixel based on its specific position. Consequently, the resulting image becomes more visually accessible to the human eye and exhibits enhanced details, albeit at the expense of longer processing times. However, these traditional operators often require manual tuning by experienced engineers, which can be cumbersome since evaluating results necessitates testing across various scenes. Although system contributions have aimed to simplify the implementation of high-performance executables~\cite{ragan2012decoupling,hegarty2014darkroom,mullapudi2016automatically,lee2016automatic}, they still necessitate programming expertise, incur runtime costs that escalate with pipeline complexity, and are only applicable when the source code for the filters is available. Therefore, seeking an automatic strategy for photo enhancement is of great interest.

\vspace{0.5em}
\noindent
\textbf{Learning-based photo enhancement.}
The learning-based approach aims to train an image enhancement model using the input and target image datasets. Bychkovsky \emph{et~al.}~\cite{bychkovsky2011learning} collected the Adobe FiveK dataset, which is the first large benchmark in the domain, and learned a mapping curve to approximate the photographer's adjustments. Some approaches \cite{afifi2020deep, chen2018deep, kim2020pienet, wang2019underexposed, ignatovweakly, deng2018aesthetic} consider image enhancement as an image translation task and design end-to-end networks to directly predict enhancement results. Deng \emph{et~al.}~\cite{deng2018aesthetic} implemented image enhancement through adversarial learning from an aesthetic perspective, Afifi \emph{et~al.}~\cite{afifi2020deep} and Kim \emph{et~al.}~\cite{kim2020pienet} designed network structures based on Unet networks to generate enhanced images, and Chen \emph{et~al.}~\cite{chen2018deep} proposed a two-way GANs to achieve unpaired image enhancement. Other approaches\cite{yan2014learning,park2018distort,afifi2019color,hu2018exposure,kosugi2020unpaired,gharbi2017deep,he2020conditional,kim2020global,li2020flexible,liu2020color,moran2020deeplpf,moran2021curl,kim2016accurate,tai2017image} combine the domain prior knowledge contained in various manual design models with neural networks. Kim \emph{et~al.}~\cite{kim2020global}, Moran \emph{et~al.}~\cite{moran2021curl} and Li \emph{et~al.}~\cite{li2020flexible} by predicting a 1D rgb curve, while Park~\emph{et~al.}~\cite{park2018distort}, Kosugi~\emph{et~al.}~\cite{kosugi2020unpaired} and Hu~\emph{et~al.}~\cite{hu2018exposure} used a reinforcement learning strategy for image enhancement, Where Hu \emph{et~al.}~\cite{hu2018exposure} trained a neural network to fit a mapping curve which can be considered as a 1D LUT for photo enhancement. Some approaches \cite{ignatov2017dslr, chen2018deep, kim2020pienet, wang2019underexposed, ignatovweakly, deng2018aesthetic, zhang2019residual} consider image enhancement as an image translation task and design end-to-end networks to directly predict enhancement results. Ignatov~\emph{et~al.}~\cite{ignatov2017dslr} constructed the dataset DPED, and used GAN to learn the mapping relationships of photo pairs. Deng \emph{et~al.}~\cite{deng2018aesthetic} implemented image enhancement through adversarial learning from an aesthetic perspective. Afifi \emph{et~al.}~\cite{afifi2020deep} and Kim \emph{et~al.}\cite{kim2020pienet} designed network structures based on Unet networks to generate enhanced images, and Chen \emph{et~al.}~\cite{chen2018deep} proposed a two-way GANs to achieve unpaired image enhancement. Recently, researchers have tried to use transformer for image enhancement~\cite{chen2021pre,dong2022incremental, liang2021swinir,wang2022multi,wang2022uformer}. Chen \emph{et~al.}~\cite{chen2021pre} provides a large pre-trained transformer for a variety of image enhancement tasks. Wang \emph{et~al.}~\cite{wang2022uformer} proposes a new structure combining unet and transformer.

\vspace{0.5em}
\noindent
\textbf{3D-LUT based method.}
Zeng \emph{et~al.}~\cite{zeng2020learning} first proposed to combine 3D LUTs with deep learning to construct a 3D-LUT based adaptive image enhancement network that learns multiple base 3D LUTs and a small convolutional neural network (CNN) simultaneously in an end-to-end manner. For each input image, an adaptive 3D LUT can be generated by fusing the base 3D LUT according to the image content to efficiently transform the color and tone of the source image. Some researchers have conducted further studies~\cite{wang2021real,cong2022high,zhang2022clut,yang2022adaint} based on~\cite{zeng2020learning}. Wang \emph{et~al.}~\cite{wang2021real} designed a lightweight two-head weight predictor with two weight outputs and learned the spatially-aware 3D LUTs that are fused in an end-to-end manner via the predicted weights. Cong \emph{et~al.}~\cite{cong2022high} proposed CDTNet with a 3D-LUT-based sub-module embedded in the network for high-resolution image coordination tasks. The use of multiple 3D LUTs leads to the increase of parameters. Zhang \emph{et~al.}~\cite{zhang2022clut} proposed an effective Compressed representation of 3-dimensional LookUp Table (CLUT) that maintains the 3D-LUT mapping capability and reduces the parameters. AdaInt~\cite{yang2022adaint} adaptively predicted the distribution of sparse sampling points in the 3D-LUT based on the input image content, improving the local nonlinear transformation capability of the 3D LUT.
To adopt LUTs for extremely efficient edge inference, Yang~\emph{et~al.}~\cite{yang2024taming} developed a network without any convolutional natural network, leveraging only pointwise convolution to extract color information.
Additionally, Kim~\emph{et~al.}~\cite{kim2024image} combined 3D LUTs and bilateral grids to deal with local differences within images.
Besides, Kosugi~\cite{kosugi2024prompt} introduced a prompt loss to guide the network training process.

Despite these advancements, existing 3D-LUT-based methods face challenges in generating high-quality details due to two key limitations. First, the global 3D-LUT operator lacks sensitivity to local features. Second, for high-resolution images, the downsampling process substantially reduces fine details. To address these issues, we propose LLF-LUT++, which incorporates a Laplacian pyramid to preserve local details while achieving efficient processing of 4K resolution images in just 13 ms.

\section{Proposed Method}
\label{method}

\begin{figure*}
\centering
\includegraphics[width=\textwidth]{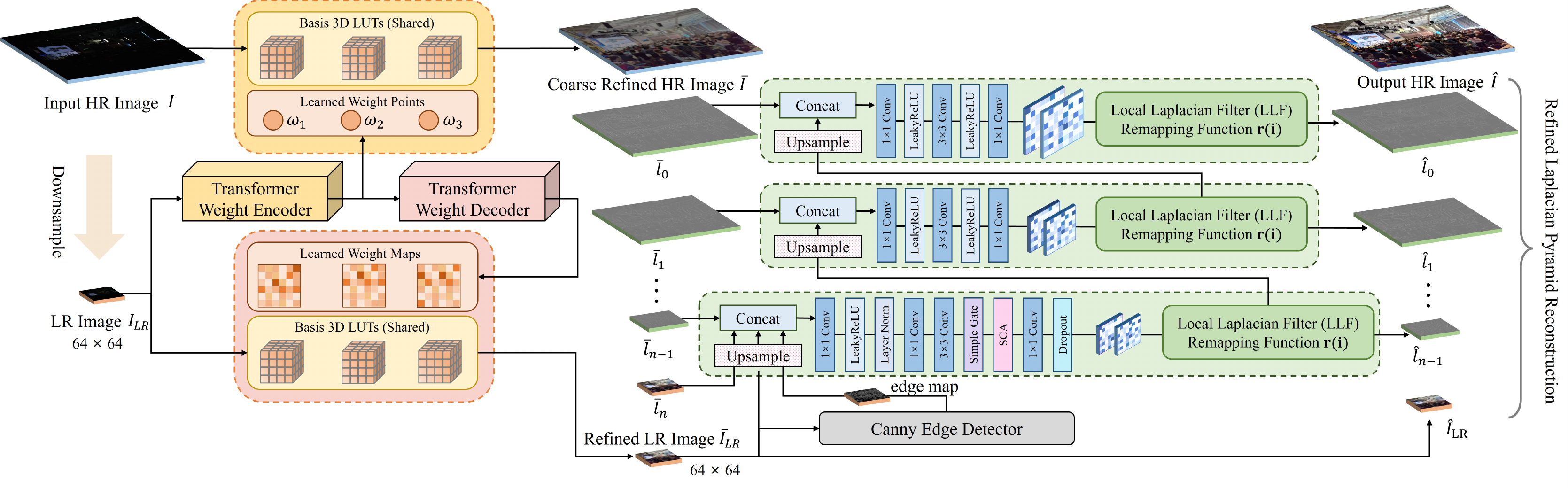}
\vspace{-2em}
\caption{
The framework of LLF-LUT++. For global enhancement, the LR image $I_{LR}$ is first fed into a lightweight transformer weight predictor to predict weight points and weight maps for refining HR image $I$ and LR image $I_{LR}$, respectively. Then the coarse refined HR image $\overline{I}$ is decomposed into a Laplacian pyramid. To adaptively refine the high-frequency components, we progressively learn an image-adaptive local Laplacian filter (LLF) based on both high- and low-frequency images. Then, we perform the remapping function of the local Laplacian filter to refine the high-frequency components while preserving the pyramid reconstruction capability. For the level $n-1$, we concatenate the component with the refined LR image $\overline{I}_{LR}$ and its edge map to mitigate potential halo artifacts.
}
\vspace{-1em}
\label{fig:arc}
\end{figure*}

We show the framework of LLF-LUT++ in Fig.~\ref{fig:arc}. We design a fusion strategy to perform an initial global enhancement using a learnable 3D lookup table and develop a transformer model to extract the global information for weight prediction. Then, that feeds the globally enhanced image into a pyramid-structured Laplacian filter for local edge enhancement to obtain the final enhanced image.
In what follows, we first explain the overview of the framework, and then we detail the key components, basis 3D LUTs fusion strategy, spatial-frequency transformer weight predictor, image-adaptive learnable local Laplacian filter and overall training objective.

\subsection{Framework Overview}
\label{overview}
We propose an end-to-end framework to manipulate tone while preserving local edge detail in photo enhancement tasks. The pipeline of our proposed method is illustrated in Fig.~\ref{fig:arc}.
Overall, we build an adaptive Laplacian pyramid network, which employs a dynamic adjustment of the decomposition levels to match the resolution of the input image. This adaptive process ensures that the low-frequency image achieves a proximity of approximately $64\times64$ resolution. The described decomposition process possesses invertibility, allowing the original image to be reconstructed by incremental operations. According to Burt and Adelson~\cite{burt1987Laplacian}, each pixel in the low-frequency image is averaged over adjacent pixels by means of an octave Gaussian filter, which reflects the global characteristics of the input image, including color and illumination attributes. Meanwhile, other high-frequency components contain edge-detailed textures of the image. Motivation by the characteristics of the Laplacian pyramid, we propose to manipulate tone on the low-frequency components while refining the high-frequency components progressively to preserve local edge details.

To manipulate tone, we design a fusion strategy by adopting weight map and weight point fusion for the HR inputs and their LR counterparts, respectively. This design integrates the benefits of weight point fusion and weight map fusion. Specifically, we first downsample the input HR image $I$ into its low-resolution $I_{LR}$. Then, with $I_{LR}$ as input, we introduce a lightweight spatial-frequency transformer for weight prediction, predicting the learned weight points and learned weight maps. These weights are employed to fuse the basis of 3D LUTs to refine the HR and LR images, resulting in $\overline{I}$ and $\overline{I}_{LR}$. This strategy achieves high performance of weight map fusion for LR images while producing rough but efficient enhancement for HR images through weight point fusion. In the output Laplacian pyramid, the $\overline{I}_{LR}$ is utilized as the low-frequency component $\hat{I}_{LR}$.

In addition, we progressively refine the higher-resolution component conditioned on the lower-resolution one. The proposed framework consists of three parts.
Firstly, given the refined HR image $\overline{I}$, we initially decompose it into an adaptive Laplacian pyramid, resulting in a collection represented by ${L}=[\overline{l}_{0},\overline{l}_{1},\cdots,\overline{l}_{n}]$, from high-frequency to low-frequency component. Here, $n$ represents the number of decomposition levels in the Laplacian pyramid. In this work, we set the lowest resolution of the pyramid to $64\times64$ for the input of 480p and 4K resolution. The downsampling factor between adjacent layers in the pyramid is $1/2$. 
Secondly, in the level $l = n-1$, we construct parameter value maps by leveraging a learned model on the concatenation of [$\overline{l}_{n-1}, up(\overline{l}_{n}), up(\overline{I}_{LR}), up(edge(\overline{I}_{LR})$], where $up(\cdots)$ represents a bilinear up-sampling operation and $edge(\cdots)$ denotes for the canny edge detector. These parameter value maps are then employed to perform a fast local Laplacian filter~\cite{aubry2014fast} on the Laplacian layer of level $n-1$. This step effectively refines the high-frequency components while considering the local edge detail information.
Lastly, we propose an efficient and progressive upsampling strategy to further enhance the refinement of the remaining Laplacian layers with higher resolutions. Starting from level $l = n-2$ down to $l = 0$, we sequentially upsample the refined components from the previous level and concatenate them with the corresponding Laplacian layer. Subsequently, we employ a lightweight convolution block to perform another fast local Laplacian filter. This iterative process iterates across multiple levels, effectively refining the high-resolution components.
In the following, we introduce the Basis 3D LUTs Fusion Strategy in Sec.~\ref{sec:fusion} and Spatial-frequency Transformer Weight Predictor in Sec.~\ref{sec:transformer}. Subsequently, we present Image-adaptive Learnable Local Laplacian Filter in Sec.~\ref{sec:filter} and the overall training objective in Sec.~\ref{sec:loss}.

\begin{figure*}
\centering
\includegraphics[width=0.9\textwidth]{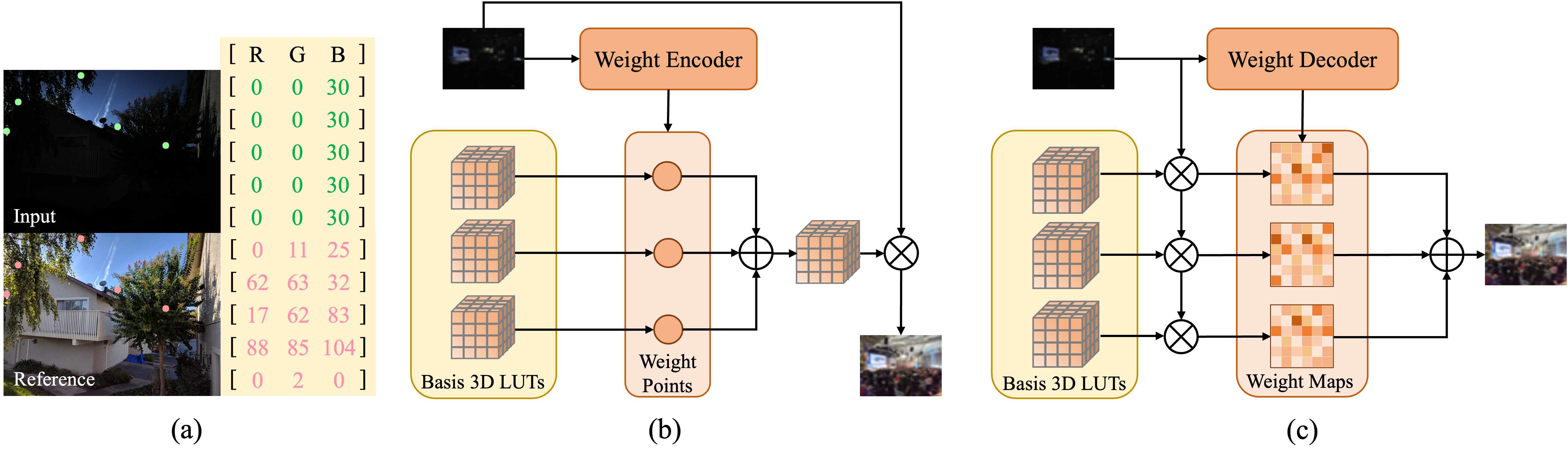}
\vspace{-1em}
\caption{Illustration of the basis 3D LUTs fusion strategy. (a) present the multiple pixel mapping relationships of an image pair; (b) is the conventional basis 3D LUTs fusion strategy; (c) is the pixel-level basis 3D LUTs fusion strategy.}
\vspace{-1em}
\label{fig:map}
\end{figure*}

\subsection{Basis 3D LUTs Fusion Strategy}
\label{sec:fusion}
According to the inherent properties of the Laplacian pyramid, the low-frequency image contains properties such as color and illumination of the images. Therefore, we employ 3D LUTs to perform tone manipulation on low-frequency images $\mathbf{I}_{LR}$. In RGB color space, a 3D LUT defines a 3D lattice that consists of $N^{3}_{b}$ elements, where $N_{b}$ is the number of bins in each color channel. Each element defines a pixel-to-pixel mapping function $\mathbf{M}^{c}(i,j,k)$, where $i,j,k=0,1,\cdots,n_{b}-1\in\mathbb{I}^{N_{b}-1}_{0}$ are elements' coordinates within 3D lattice and $c$ indicates color channel. Given an input RGB color \{(${I}^{r}_{(i,j,k)},{I}^{g}_{(i,j,k)},{I}^{b}_{(i,j,k)}$)\}, where $i,j,k$ are indexed by the corresponding RGB value, a output ${O}^{c}$ is derived by the mapping function as follows:
\begin{equation}
{O}^{c}_{(i,j,k)}=\mathbf{M}^{c}({I}^{r}_{(i,j,k)},{I}^{g}_{(i,j,k)},{I}^{b}_{(i,j,k)}) .
\label{eq:lut}
\end{equation}

The mapping capabilities of conventional 3D LUTs are inherently constrained to fixed transformations of pixel values. Fig.~\ref{fig:map}(a) demonstrates this limitation, where the input image has the same pixel values at different locations. However, these locations contain different pixel values in the reference image. While the input image is interpolated through a look-up table, the transformed image retains the same transformed pixel values at these locations. Consequently, the conventional 3D LUT framework fails to accommodate intricate pixel mapping relationships, thus impeding its efficacy in accurately representing such pixel transformations.

Inspired by~\cite{wang2021real}, we propose an effective 3D LUT fusion strategy to address this inherent limitation. The conventional 3D LUT fusion strategy proposed by~\cite{zeng2020learning} is shown in Fig.~\ref{fig:map}(b), which first utilizes the predicted weights to fuse the multiple 3D LUTs into an image-adaptive one and then performs trilinear interpolation to transform images. In contrast, as shown in Fig.~\ref{fig:map}(c), our strategy is first to perform trilinear interpolation with each LUT and then fuse the enhanced image with predicted pixel-level weight maps. In this way, our method can enable a relatively more comprehensive and accurate representation of the complex pixel mapping relationships through the weight values of each pixel. The pixel-level mapping function ${\Phi}^{h,w,c}$ can be described as follows:
\begin{equation}
\begin{split}
{O}^{h,w,c}_{(i,j,k)}={\Phi}^{h,w,c}({I}^{r}_{(i,j,k)},{I}^{g}_{(i,j,k)},{I}^{b}_{(i,j,k)},{\omega}^{h,w}) \\ = \sum_{t=0}^{T-1} {\omega}^{h,w}_{t} {M}^{c}_{t}({I}^{r}_{(i,j,k)},{I}^{g}_{(i,j,k)},{I}^{b}_{(i,j,k)}) ,
\end{split}
\label{eq:slut}
\end{equation}
\noindent where ${O}^{h,w,c}_{(i,j,k)}$ is the final pixel-level output, ${\omega}^{h,w}_{t}$ represents a pixel-level weight map for $t$-th 3D LUTs located at $(h,w)$. Note that our proposed strategy involves the utilization of multiple trilinear interpolations, which may impact the computational speed when applied to high-resolution images. However, since our method operates at the resolution of $64\times64$, the computational overhead is insignificant.

As shown in Fig.~\ref{fig:arc}, given ${I}_{LR}$ with a reduced resolution, we feed it into a weight predictor to output the content-dependent weight maps. Moreover, to integrate the benefits of weight point fusion and weight map fusion, we generate additional learned weight points to refine the HR image $I$. It retains the high performance of weight map fusion and achieves rough but efficient enhancement for HR images through weight point fusion. Notably, the weight points and weight maps are predicted simultaneously by the same transformer network, which ensures enhancement consistency between HR and LR images. The refined LR image $\overline{I}_{LR}$ is utilized as the lowest frequency component $\hat{I}_{LR}$ of the refined Laplacian pyramid to reconstruct output.

\begin{figure*}
\begin{center}
\includegraphics[width=\textwidth]{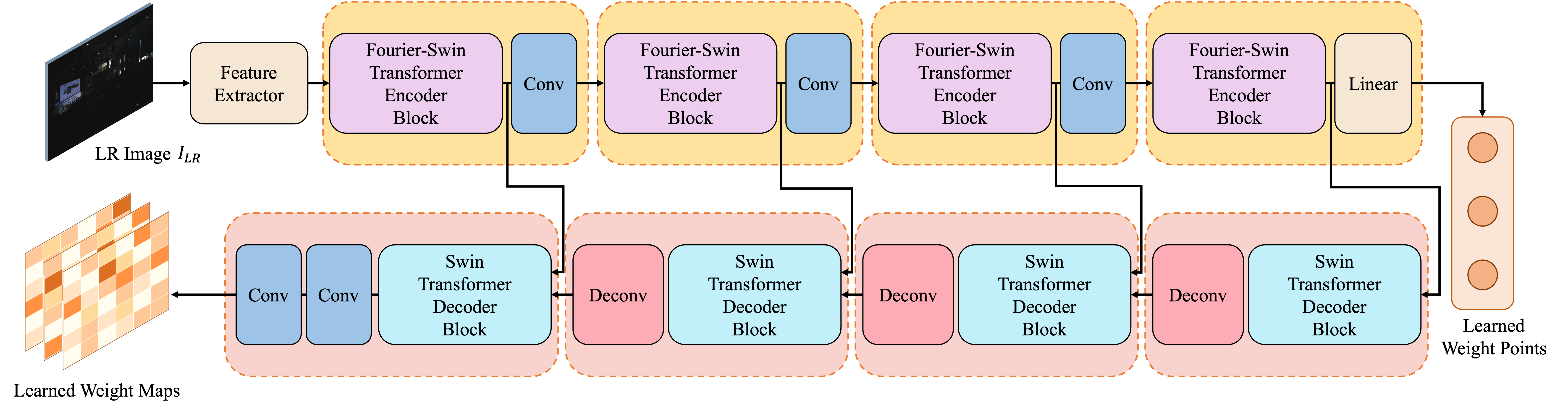}
\end{center}
\vspace{-1.5em}
\caption{The architecture of the proposed spatial-frequency transformer network for weight prediction. The features extracted from input LR images $I_{LR}$ are processed by four encoders to build dictionaries that will be used as inputs for the decoders. The last encoder block will output learnable weight points through a linear layer, while the last decoder block will output learnable weight maps through two convolutional layers. The Fourier-Swin transformer encoder and Swin transformer decoder blocks are illustrated with more details in Fig.~\ref{fig:enc-dec}.}
\vspace{-0.5em}
\label{fig:trans}
\end{figure*}

\subsection{Spatial-frequency Transformer Weight Predictor}
\label{sec:transformer}

\begin{figure}
\begin{center}
\includegraphics[width=0.48\textwidth]{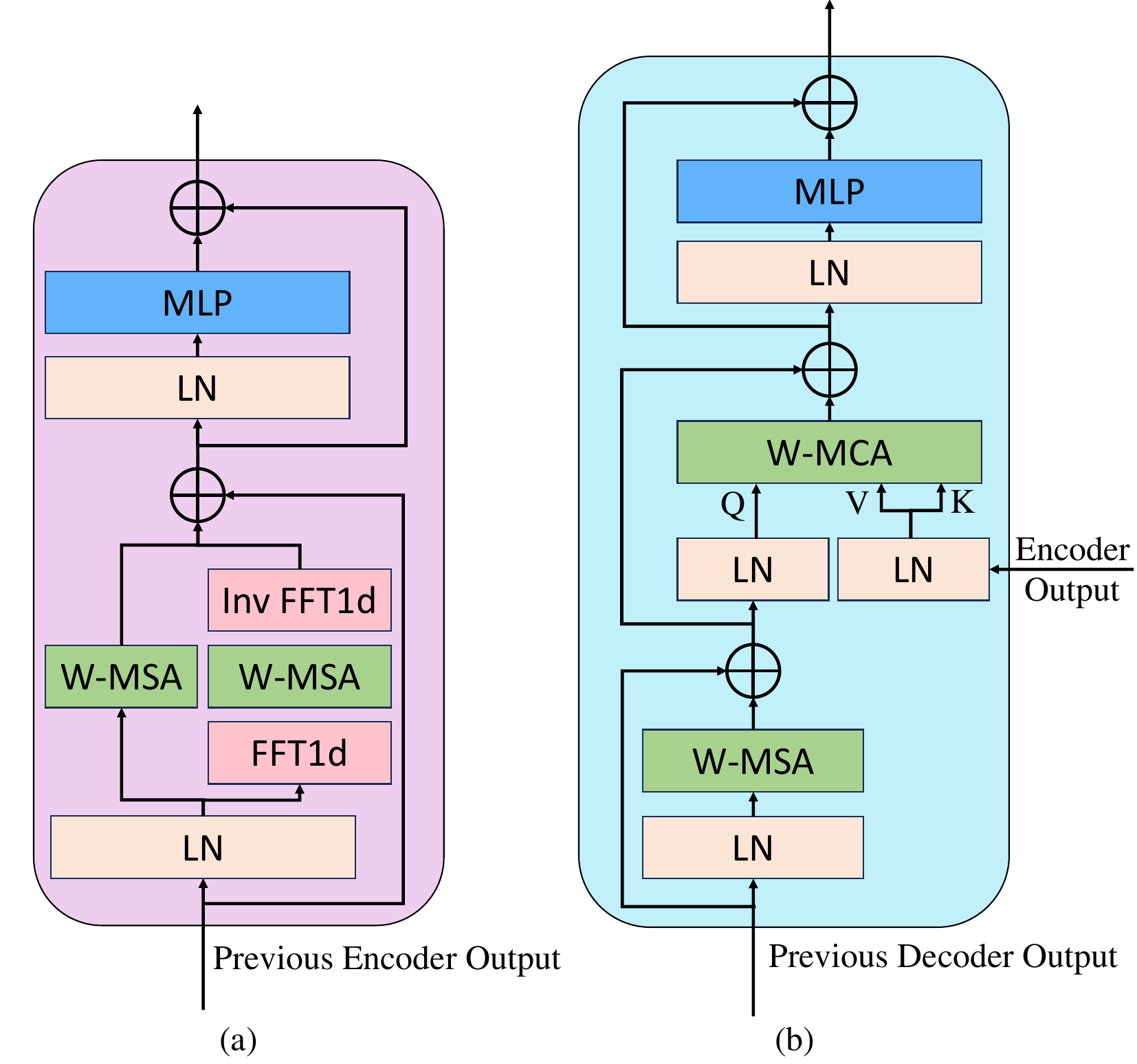}
\end{center}
\vspace{-1em}
\caption{(a) is the Fourier-Swin Transformer encoder block; (b) is the basic Swin Transformer decoder block. Here, LN stands for Layer Normalization, W-MSA is Windowed Multi-Head Self-Attention, and W-MCA is Windowed Multi-Head Cross-Attention. FFT1d and inv FFT1d indicate the Fast Fourier Transformer and its inverse.}
\vspace{-0.5em}
\label{fig:enc-dec}
\end{figure}

Given that the weight predictor aims to capture global context, including brightness, color, and tone of an image, a transformer backbone is more effective for extracting global information than a CNN backbone. 
However, the inefficiency and non-specific design of the existing transformer backbone presents a challenge in enabling the weight prediction network to output both weight maps and weight points. To address this issue, we proposed a UNet-style spatial-frequency transformer network for weight prediction, as depicted in Fig.~\ref{fig:trans}. This design not only shortens runtime but also enhances performance.

{The feature extractor comprises a 3×3 convolutional layer followed by a LeakyReLU activation layer, designed to capture discriminative feature representations from the input image, which are subsequently propagated to the encoder for further processing.}

The encoder comprises four encoder blocks, each incorporating a Fourier-Swin Transformer encoder block, as illustrated in Fig.~\ref{fig:enc-dec} (a). Unlike the vanilla Swin-Transformer block~\cite{liu2021swin}, our design features two branches to investigate spatial and spectral information separately. Specifically, we employ Windows Multi-Head Self-Attention (W-MSA) to extract spatial features and introduce the Fast Fourier Transform (FFT) and its inverse to W-MSA for spectral domain perception. The two features are then combined, utilizing a simple yet effective fusion strategy. By leveraging both spatial and spectral information, the encoder can better extract high-frequency detail information while maintaining negligible computational overhead due to the low resolution of the downsampled input image. This transformer network design, with significantly reduced resolution, saves substantial runtime compared to the previous GRL transformer~\cite{li2023efficient} and further shortens runtime by reducing the number of transformer blocks. Additionally, the first three encoder blocks each cascade a convolutional layer for image downsampling, while the final encoder block employs a linear layer to output three weight points. These weight points are used for fusing the basis 3D-LUT and enhancing the input image $I$, resulting in a coarsely refined image $\bar{I}$.

Similarly, the decoder comprises four decoder blocks. The first three decoder blocks each include a Swin-Transformer decoder block and a transposed convolution layer, while the last decoder block consists of a Swin-Transformer decoder block and two convolutional layers, predicting three weight maps. These weight maps are used for fusing the basis 3D-LUT and enhancing the input LR image $I_{LR}$, resulting in a coarsely refined LR image $\bar{I}_{LR}$. As shown in Fig.~\ref{fig:enc-dec} (b), to fuse the decoder output and skip encoder output, each Swin-Transformer decoder block interpolates a Windowed Multi-Head Cross-Attention (W-MCA)~\cite{geng2022rstt} into a Swin-Transformer block. Specifically, in W-MCA, the query is generated from the decoder output, while the key and value use the skip encoder output.

\subsection{Image-adaptive Learnable Local Laplacian Filter}
\label{sec:filter}

Although the basis 3D LUTs fusion strategy demonstrates stable and efficient enhancement of input images across various scenes, the transformation of pixel values through weight maps alone still falls short of significantly improving local detail and contrast. To tackle this limitation, one potential solution is to integrate a local enhancement method with 3D LUT. In this regard, drawing inspiration from the intrinsic characteristics of the Laplacian pyramid~\cite{burt1987Laplacian}, which involves texture separation, visual attribute separation, and reversible reconstruction, the combination of 3D LUT and the local Laplacian filter~\cite{paris2011local} can offer substantial benefits.

Local Laplacian filters are edge-aware local photo enhancement operators that define the output image by constructing its Laplacian pyramid coefficient by coefficient. The computation of each coefficient $i$ is independent of the others. These coefficients are computed by the following remapping functions $\mathbf{r(i)}$:
\begin{equation}
\mathbf{r(i)}=\left\{
\begin{array}{lc}
g+sign(i-g)\sigma_{r}(|i-g|/\sigma_{r})^{\alpha} & if \: i \leq \sigma_{r} \\
g+sign(i-g)(\beta(|i-g|-\sigma_{r})+\sigma_{r}) & if \: i > \sigma_{r} 
\end{array} \right. ,
\label{eq:remap}
\end{equation}
\noindent where $g$ is the coefficient of the Gaussian pyramid at each level, which acts as a reference value, $sign(x)=x/|x|$ is a function that returns the sign of a real number, $\alpha$ is one parameter that controls the amount of detail increase or decrease, $\beta$ is another parameter that controls the dynamic range compression or expansion, and $\sigma_{r}$ defines the intensity threshold the separates details from edges.

Nevertheless, the conventional approach described in Eq.~\ref{eq:remap} necessitates manual parameter adjustment for each input image, leading to a cumbersome and labor-intensive process. To overcome this limitation, we propose an image-adaptive learnable local Laplacian filter (LLF) to learn the parameter value maps for the remapping function. The objective function of the learning scheme can be written as follows:
\begin{equation}
min_{\alpha,\beta} \mathcal{L}(\mathbf{r}(l,g),{R}) ,
\label{eq:objective}
\end{equation}
\noindent where $\alpha$ and $\beta$ are the learned parameter value maps of the Laplacian pyramid, $\mathcal{L}(\cdots)$ denotes the loss functions, $\mathbf{r}(l,g)$ presents the image-adaptive learnable local Laplacian filter (LLF), $l$ and $g$ are the coefficients of the Laplacian and Gaussian pyramid, respectively, ${R}$ is the reference image. Note that the parameter $\sigma_{r}$ does not impact the filter's performance; thus, it is fixed at 0.1 in this paper. Furthermore, to enhance computational efficiency, we have employed the fast local Laplacian filter~\cite{aubry2014fast} instead of the conventional local Laplacian filter.

As discussed in Sec.~\ref{overview}, we have $\overline{l}_{n-1}\in\mathbb{R}^{\frac{h}{2^{n-1}}\times \frac{w}{2^{n-1}}\times3}$ and $\overline{l}_{n},\overline{I}_{LR}\in\mathbb{R}^{\frac{h}{2^{n}}\times \frac{w}{2^{n}}\times3}$. To address potential halo artifacts, we initially employ a Canny edge detector with default parameters to extract the edge map of $\overline{I}_{LR}$. Subsequently, we upsample $\overline{l}_{n}$, $\overline{I}_{LR}$ and $edge(\overline{I}_{LR})$ using bilinear operations to match the resolution of $l_{N-1}$ and concatenate them.
The concatenated components are fed into a line of neural network layers to generate parameter $\alpha_{n-1},\beta_{n-1}$ as depicted in Fig.~\ref{fig:arc}.
The outputs are utilized for the remapping function $\mathbf{r(i)}$ to refine $\overline{l}_{n-1}$:
\begin{equation}
\hat{l}_{n-1}=\mathbf{r}(\overline{l}_{n-1},g_{n-1},\alpha_{n-1},\beta_{n-1}) .
\label{eq:refine}
\end{equation}

Subsequently, we adopt a progressive upsampling strategy to match the refined high-frequency component $\hat{l}_{n-1}$ with the remaining high-frequency components. This upsampled component is concatenated with $\overline{l}_{n-2}$. As depicted in Fig.~\ref{fig:arc}, the concatenated vector $[\overline{l}_{n-2},up(\hat{l}_{n-1})]$ is feed into another LLF. The refinement process continues iteratively, progressively upsampling until $\hat{l}_{0}$ is obtained. By applying the same operations as described in Eq.~\ref{eq:refine}, all high-frequency components are effectively refined, leading to a set of refined components $[\hat{l}_{0},\hat{l}_{1},\ldots,\hat{l}_{n-1}]$. Finally, the result image $\hat{I}$ is reconstructed using the refined LR image $\hat{I}_{LR}$ with refined components $[\hat{l}_{0},\hat{l}_{1},\ldots,\hat{l}_{n-1}]$.

However, the adaptive Laplacian pyramid network in our previous confluence version LLF-LUT employs the same network structure for each layer, leading to mediocre performance. Implementing a more complex network structure at the lowest layer can improve the understanding of low-frequency information and significantly enhance performance with only a slight increase in computational cost. Additionally, the previous structure failed to effectively utilize the Gaussian layer information of the input image. Incorporating this information will enable the network to learn effective features more rapidly.

To address these issues, as illustrated in Fig.~\ref{fig:arc}, we enhanced the perception capability of the Laplacian pyramid by improving both the network input and structure used for predicting Laplacian filter parameters in the following ways: 1) We incorporated each layer of the Gaussian pyramid as network input, enabling the network to more accurately learn filter parameters by integrating color, texture, and other detailed information. 2) We redesigned the parameter learning process for the lowest layer by incorporating channel attention, thereby enhancing the model's interaction with information across channels. This modification allows for more effective extraction of information from the pyramid layers. 3) We added an additional convolutional layer at the lowest layer to deepen the network, facilitating more complex information aggregation. To balance the increased parameters, we replaced 3x3 convolutional layers with 1x1 convolutional layers in the subsequent network structures of the Laplacian pyramid, thereby reducing the overall parameter count.

{To enhance the performance of the network, we incorporated key design elements from the NAF-Block~\cite{chen2022simple} into our architecture, including the Simplified Channel Attention (SCA) module, the Simple Gate mechanism, and dropout layers. These components collectively contribute to the improved performance of the proposed model.}

\subsection{Overall Training Objective}
\label{sec:loss}

The proposed framework is trained in a supervised scenario by optimizing a reconstruction loss. To encourage a faithful global and local enhancement, given a set of image pairs $({I},{R})$, where ${I}[i]$ and ${R}[i]$ denote a pixel pair of 16-bit input HDR and 8-bit reference LDR image, we define the reconstruction loss function as follows:
\begin{equation}
    \mathcal{L}_{1} = \sum_{i=1}^{H\times W} (\parallel \hat{{I}}[i] - {R}[i] \parallel_1 + \parallel \hat{{I}}_{LR}[i] - {R}_{LR}[i] \parallel_1) ,
\end{equation}
\noindent where $\hat{{I}}[i]$ is the output of network with ${I}[i]$ as input, $\hat{{I}}_{LR}[i]$ is the output of 3D LUT with ${I}_{LR}[i]$ as input, ${R}_{LR}[i]$ is the low-frequency image of reference image ${R}[i]$.

To make the learned 3D LUTs more stable and robust, some regularization terms from~\cite{zeng2020learning}, including smoothness term $\mathcal{L}_{s}$ and monotonicity term $\mathcal{L}_{m}$, are employed. In addition to these terms, we employ an LPIPS loss~\cite{zhang2018unreasonable} function that assesses a solution concerning perceptually relevant characteristics (\eg, the structural contents and detailed textures):
\begin{equation}
\mathcal{L}_{p} = \sum_{l} \frac{1}{H^{l}W^{l}} \sum_{h,w} \parallel \phi(\hat{{I}})^{l}_{hw} - \phi({R})^{l}_{hw} \parallel_2^2 ,
\end{equation}
\noindent where $\phi(\cdot)^{l}_{hw}$ denotes the feature map of layer $l$ extracted from a pre-trained AlexNet~\cite{krizhevsky2017imagenet}.

To summarize, the complete objective of our proposed model is combined as follows:
\begin{equation}
\mathcal{L} = \mathcal{L}_{1} + \lambda_{s}\mathcal{L}_{s} + \lambda_{m}\mathcal{L}_{m} + \lambda_{p}\mathcal{L}_{p},
\end{equation}
\noindent where $\lambda_{s}$, $\lambda_{m}$, and $\lambda_{p}$ are hyper-parameters to control the balance of loss functions. In our experiment, these parameters are set to $\lambda_{s}= 0.0001$, $\lambda_{m}= 10$, $\lambda_{p}= 0.01$.

\begin{table*}
    \caption{Quantitative comparison on HDR+~\cite{hasinoff2016burst} dataset. The Multiply–Accumulate Operations (MACs) and runtime are both computed on $3840\times2160$ resolution. ``N.A." means that the results are not available due to insufficient memory of the GPU.}
    \vspace{-0.5em}
    \centering
    \setlength{\tabcolsep}{7.5pt}
    \begin{tabular}{l | c c c | c c c c || c c c c}
        \toprule[0.15em]
        \multirow{2}{*}{Methods} &\multirow{2}{*}{$\#$Params} & \multirow{2}{*}{MACs}  & \multirow{2}{*}{Runtime}  &\multicolumn{4}{c||}{HDR+ (480p)} &\multicolumn{4}{c}{HDR+ (original)} \\
        \cmidrule(r){5-8} \cmidrule(r){9-12}
        & & & &PSNR$\textcolor{black}{\uparrow}$ &SSIM$\textcolor{black}{\uparrow}$ &LPIPS$\textcolor{black}{\downarrow}$ &$\triangle E$$\textcolor{black}{\downarrow}$ &PSNR$\textcolor{black}{\uparrow}$ &SSIM$\textcolor{black}{\uparrow}$ &LPIPS$\textcolor{black}{\downarrow}$ &$\triangle E$$\textcolor{black}{\downarrow}$ \\
        \midrule[0.15em]
        UPE~\cite{wang2019underexposed} &999K & 1.146G & 8.42ms &23.33 &0.852 &0.150 &7.68 &21.54 &0.723 &0.361 &9.88      \\
        HDRNet~\cite{gharbi2017deep} &482K & 1.103G & 4.56ms &24.15 &0.845 &0.110 &7.15 &23.94 &0.796 &0.266 &6.77      \\
        CSRNet~\cite{he2020conditional} &37K & 52.84G & 48.28ms &23.72 &0.864 &0.104 &6.67 &22.54 &0.766 &0.284 &7.55 \\
        DeepLPF~\cite{moran2020deeplpf} &1.72M& 454.4G & 386.62ms &25.73 &0.902 &0.073 &6.05 &N.A. &N.A. &N.A. &N.A. \\
        LUT~\cite{zeng2020learning} &592K & 0.676G & 2.67ms &23.29 &0.855 &0.117 &7.16 &21.78 &0.772 &0.303 &9.45  \\
        sLUT~\cite{wang2021real} &4.52M & 30.32G & 9.63ms &26.13 &0.901 &0.069 &5.34 &23.98 &0.789 &0.242 &6.85 \\
        CLUT~\cite{zhang2022clut} &952K & 9.391G & 8.35ms &26.05 &0.892 &0.088 &5.57 &24.04 &0.789 &0.245 &6.78  \\
        {LUTwithBGrid~\cite{kim2024image}} & {464K} & {0.242G} & {4.51ms} & {22.89} & {0.844} & {0.140} & {10.22} & {21.96} & {0.749} & {0.286} & {10.54}  \\
        {PG-IA-NILUT~\cite{kosugi2024prompt}} & {460K} & {14.47G} & {242ms} & {23.52} & {0.833} & {0.121} & {9.93} & {22.64} & {0.759} & {0.295} & {9.89}  \\
        
        {DPRNet~\cite{yang2024learning}} & {2.9M} & {6.9G} & {104.6ms} & {\underline{27.85}} & {\textbf{0.927}} & {\textbf{0.036}} & {5.68} & {\underline{27.90}} & {\textbf{0.917}} & {\textbf{0.061}} & {\underline{5.32}} \\
        \midrule
        \midrule
        \textbf{{LLF-LUT}} &731K & 2.923G & 20.51ms &\underline{26.62} &{0.907} &{0.063} &\underline{5.31} &{25.32} &{0.849} &\underline{0.149} &{6.03}\\
        
        \textbf{{LLF-LUT++}} &717K & 3.720G & 13.50ms &\textbf{28.43} &\underline{0.924} & \underline{0.056} &\textbf{4.54} &\textbf{27.96} &\underline{0.882} &0.155 &\textbf{4.69} \\
        \bottomrule[0.15em]
    \end{tabular}
    \label{table:hdr}
\end{table*}

\begin{table*}
    \caption{Quantitative comparison on MIT-Adobe FiveK~\cite{bychkovsky2011learning} dataset. The Multiply–Accumulate Operations (MACs) and runtime are both computed on $3840\times2160$ resolution. ``N.A." means that the results are not available due to insufficient memory of the GPU.}
    \vspace{-0.5em}
    \setlength{\tabcolsep}{7.5pt}
    \centering
    \begin{tabular}{l | c c c | c c c c || c c c c}
        \toprule[0.15em]
        \multirow{2}{*}{Methods} &\multirow{2}{*}{$\#$Params} & \multirow{2}{*}{MACs}  & \multirow{2}{*}{Runtime}  &\multicolumn{4}{c||}{MIT-FiveK (480p)} &\multicolumn{4}{c}{MIT-FiveK (original)} \\
        \cmidrule(r){5-8} \cmidrule(r){9-12}
        & & & &PSNR$\textcolor{black}{\uparrow}$ &SSIM$\textcolor{black}{\uparrow}$ &LPIPS$\textcolor{black}{\downarrow}$ &$\triangle E$$\textcolor{black}{\downarrow}$ &PSNR$\textcolor{black}{\uparrow}$ &SSIM$\textcolor{black}{\uparrow}$ &LPIPS$\textcolor{black}{\downarrow}$ &$\triangle E$$\textcolor{black}{\downarrow}$ \\
        \midrule[0.15em]
        UPE~\cite{wang2019underexposed} &999K & 1.146G & 8.42ms &21.82 &0.839 &0.136 &9.16 &20.41 &0.789 &0.253 &10.81     \\
        HDRNet~\cite{gharbi2017deep} &482K & 1.103G & 4.56ms &23.31 &0.881 &0.075 &7.73 &22.99 &0.868 &0.122 &7.89     \\
        CSRNet~\cite{he2020conditional} &37K & 52.84G & 48.28ms  &25.31 &0.909 &\textbf{0.052} &6.17 &24.23 &0.891 &0.099 &7.10  \\
        DeepLPF~\cite{moran2020deeplpf} &1.72M & 454.4G & 386.62ms &24.97 &0.897 &0.061 &6.22 &N.A. &N.A. &N.A. &N.A.\\
        LUT~\cite{zeng2020learning} &592K & 0.676G & 2.67ms &25.10 &0.902 &0.059 &6.10 &23.27 &0.876 &0.111 &7.39 \\
        sLUT~\cite{wang2021real} &4.52M & 30.32G & 9.63ms  &24.67 &0.896 &0.059 &6.39 &24.27 &0.876 &0.103 &6.59\\
        CLUT~\cite{zhang2022clut} &952K & 9.391G & 8.35ms  &24.94 &0.898 &0.058 &6.71 &23.99 &0.874 &0.106 &7.07 \\
        {LUTwithBGrid~\cite{kim2024image}} & {464K} & {0.242G} & {4.51ms} & {\underline{25.59}} & {\textbf{0.932}} & {0.103} & {7.14} & {\underline{24.57}} &  {\textbf{0.931}} & {0.098} & {7.73} \\
        
        {PG-IA-NILUT~\cite{kosugi2024prompt}} & {460K} & {14.47G} & {242ms} & {24.31} & {0.852} & {0.062} & {10.16} & {20.86} & {0.802} & {0.140} & {11.12} \\
        
        {DPRNet~\cite{yang2024learning}} & {2.9M} & {6.9G} & {104.6ms} & {23.47} & {0.879} & {0.060} & {6.12} & {23.02} & {0.853} & {0.107} & {7.56} \\
        \midrule
        \midrule
        \textbf{LLF-LUT} &731K & 2.923G & 20.51ms &{25.53} &{0.910} &0.055 &\underline{5.64} &24.52 &\underline{0.897} &\textbf{0.081} &\textbf{6.34} \\
        
        \textbf{LLF-LUT++} &717K & 3.720G & 13.50ms &\textbf{26.06} &\underline{0.912} & \underline{0.054} &\textbf{4.93} &\textbf{24.84} &\underline{0.897} &\underline{0.086} &\underline{6.38} \\
      
        \bottomrule[0.15em]
    \end{tabular}
    \label{table:mit}
\end{table*}

\section{Experiments}
In this section, we perform quantitative and qualitative assessments of the results produced by the proposed LLF-LUT++ and other state-of-the-art methods. We first introduce the experimental setup. Then, we individually present quantitative comparison results and qualitative comparison results, followed by a runtime analysis. Moreover, we conduct several ablation studies on each component of our proposed framework, the effect of basis 3D-LUT fusion strategy and the number of Laplacian pyramid layers.

\subsection{Experimental Setup}
In this part, we introduce the used datasets, evaluation metrics, and implementation details.

\vspace{0.5em}
\noindent
\textbf{Datasets.}
We evaluate the performance of our networks on two challenging benchmark datasets: MIT-Adobe FiveK~\cite{bychkovsky2011learning} and HDR+ burst photography~\cite{hasinoff2016burst}. The MIT-Adobe FiveK dataset is widely recognized as a benchmark for evaluating photographic image adjustments. This dataset comprises 5000 raw images, each retouched by five professional photographers. In line with previous studies~\cite{zeng2020learning,wang2021real,zhang2022clut}, we utilize the ExpertC images as the reference images and adopt the same data split, with 4500 image pairs allocated for training and 500 image pairs for testing purposes. The HDR+ dataset is a burst photography dataset collected by the Google camera group to research high dynamic range (HDR) and low-light imaging on mobile cameras. We post-process the aligned and merged frames (DNG images) into 16-bit TIF images as the input and adopt the manually tuned JPG images as the corresponding reference images. We conduct experiments on both the 480p resolution and 4K resolution. The aspect ratios of source images are mostly 4:3 or 3:4.

\vspace{0.5em}
\noindent
\textbf{Evaluation metrics.}
We employ four commonly used metrics to quantitatively evaluate the enhancement performance on the datasets as mentioned above. The $\triangle E$ metric is defined based on the $L_{2}$ distance in the CIELAB color space. The PSNR and SSIM are calculated by corresponding functions in \textit{skimage.metrics} library and RGB color space. Note that higher PSNR/SSIM and lower LPIPS/$\triangle E$ indicate better performance.

\vspace{0.5em}
\noindent
\textbf{Implementation Details.}
To optimize the network, we employ the Adam optimizer~\cite{kingma2014adam} for training. The initial values of the optimizer's parameters, $\beta_{1}$ and $\beta_{2}$, are set to 0.9 and 0.999, respectively. The initial learning rate is set to $2\times10^{-4}$, and we use a batch size of 1 during training. In order to augment the data, we perform horizontal and vertical flips. The training process consists of 200 epochs. The implementation is conducted on the Pytorch~\cite{paszke2017automatic} framework with NVIDIA Tesla V100 32GB GPUs.

We comprehensively compare our proposed networks with state-of-the-art learning-based methods for enhancement in the camera-imaging pipeline. The methods included in the comparison are UPE~\cite{wang2019underexposed}, DeepLPF~\cite{moran2020deeplpf}, HDRNet~\cite{gharbi2017deep}, CSRNet~\cite{he2020conditional}, 3DLUT~\cite{zeng2020learning}, spatial-aware 3DLUT~\cite{wang2021real}, CLUT-Net~\cite{zhang2022clut}, {LUTwithBGrid~\cite{kim2024image}, PG-IA-NILUT~\cite{kosugi2024prompt}, DPRNet~\cite{yang2024learning}}, and LLF-LUT\cite{zhang2024lookup}. To simplify the notation, we use the abbreviations LUT, sLUT, and CLUT to represent 3DLUT, spatial-aware 3DLUT, and CLUT-Net, respectively, in our comparisons. It is important to note that the input images considered in our evaluation are 16-bit uncompressed images in the CIE XYZ color space, while the reference images are 8-bit compressed images in the sRGB color space.

Among the considered methods, DeepLPF and CSRNet are pixel-level methods based on ResNet and U-Net backbone, while HDRNet and UPE belong to patch-level methods, and LUT, sLUT, and CLUT are the image-level methods. Our method also falls within the image-level category. These methods are trained using publicly available source codes with recommended configurations, except for sLUT. Since the training code and weights of this method have never been released, we reproduce the results according to the description in the released article.

\begin{figure*}
    \centering
    \includegraphics[width=1\linewidth]{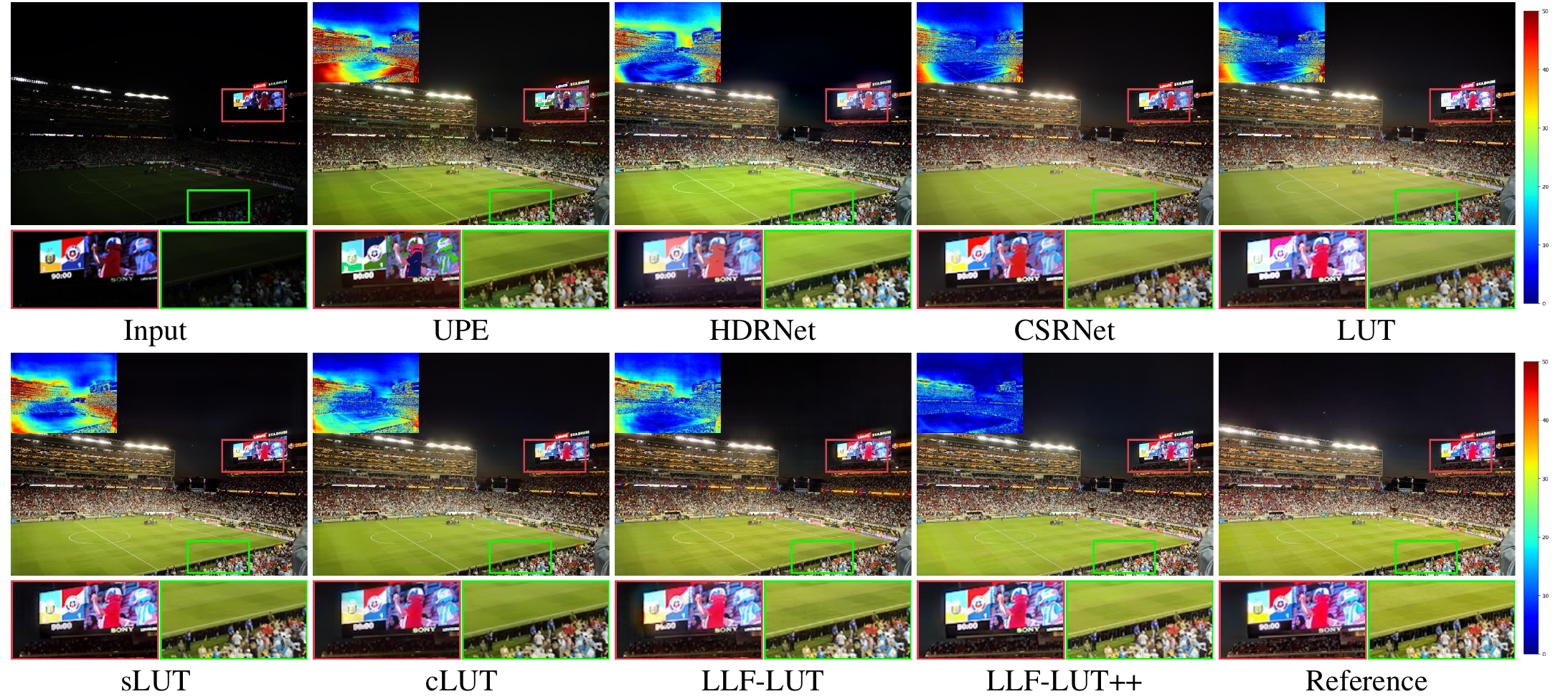}
    \vspace{-2em}
    \caption{Visual comparison with state-of-the-art methods on a 480p resolution test image from the HDR+ dataset~\cite{hasinoff2016burst}. The error maps in the upper left corner facilitate a more precise determination of performance differences. Best viewed in color and by zooming in.}    
    \label{fig:results_480p_hdr}
\end{figure*}

\begin{figure*}
    \centering
    \includegraphics[width=1\linewidth]{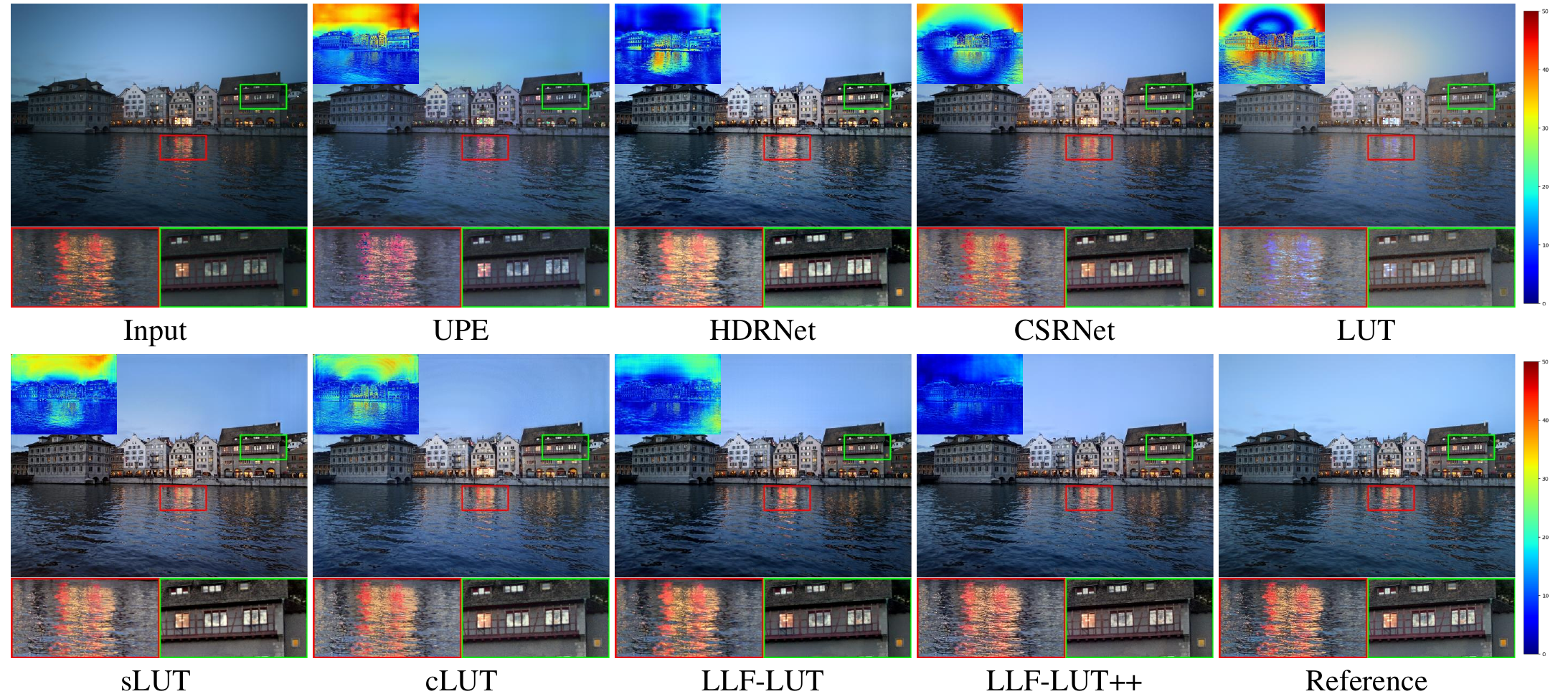}
    \vspace{-2em}
    \caption{Visual comparison with state-of-the-art methods on an original resolution test image from the HDR+ dataset~\cite{hasinoff2016burst}. The error maps in the upper left corner facilitate a more precise determination of performance differences. Best viewed in color and by zooming in.}    
    \label{fig:results_4K_hdr}
\end{figure*}

\begin{figure*}
    \centering
    \includegraphics[width=1\linewidth]{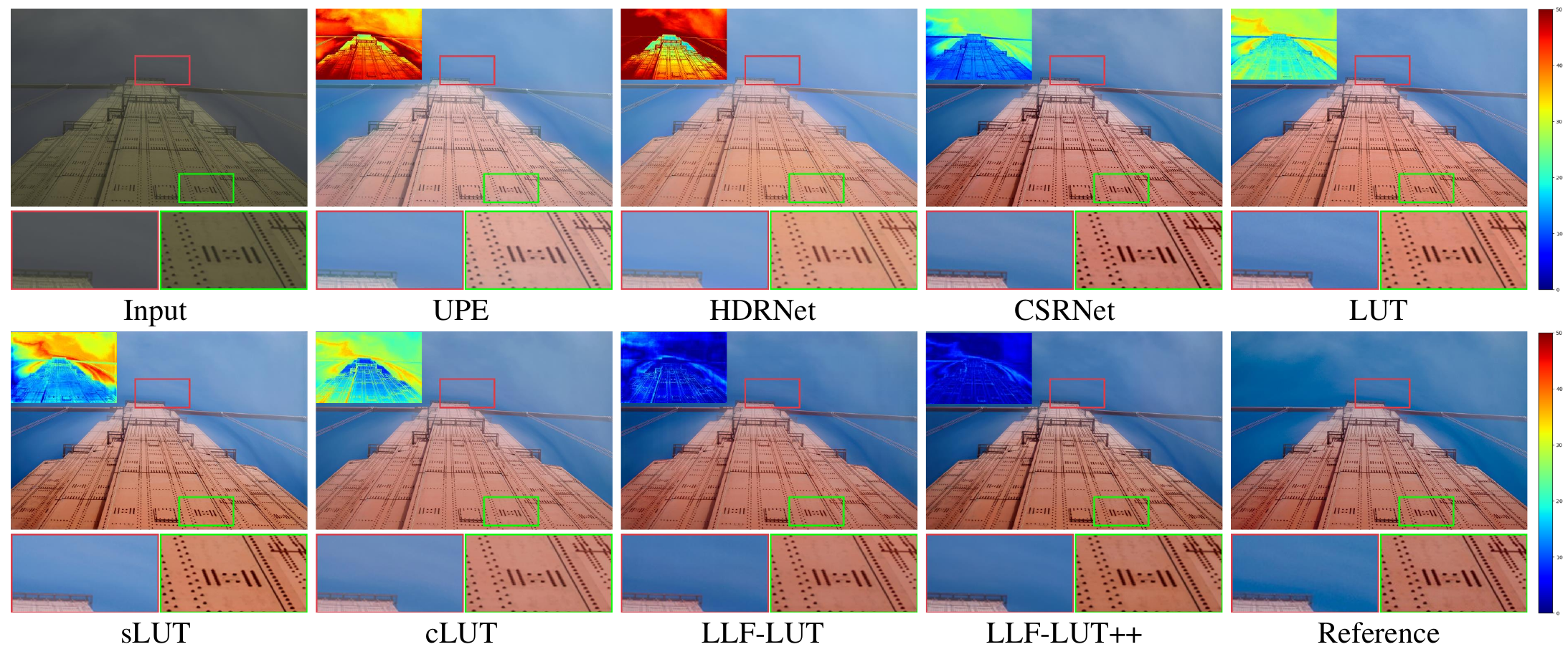}
    \vspace{-2em}
    \caption{Visual comparison with state-of-the-art methods on a 480p resolution test image from the MIT-Adobe FiveK dataset~\cite{bychkovsky2011learning}. The error maps in the upper left corner facilitate a more precise determination of performance differences. Best viewed in color and by zooming in.}
    \vspace{-0.5em}
    \label{fig:results_480p_fivek}
\end{figure*}

\begin{figure*}
    \centering
    \includegraphics[width=1\linewidth]{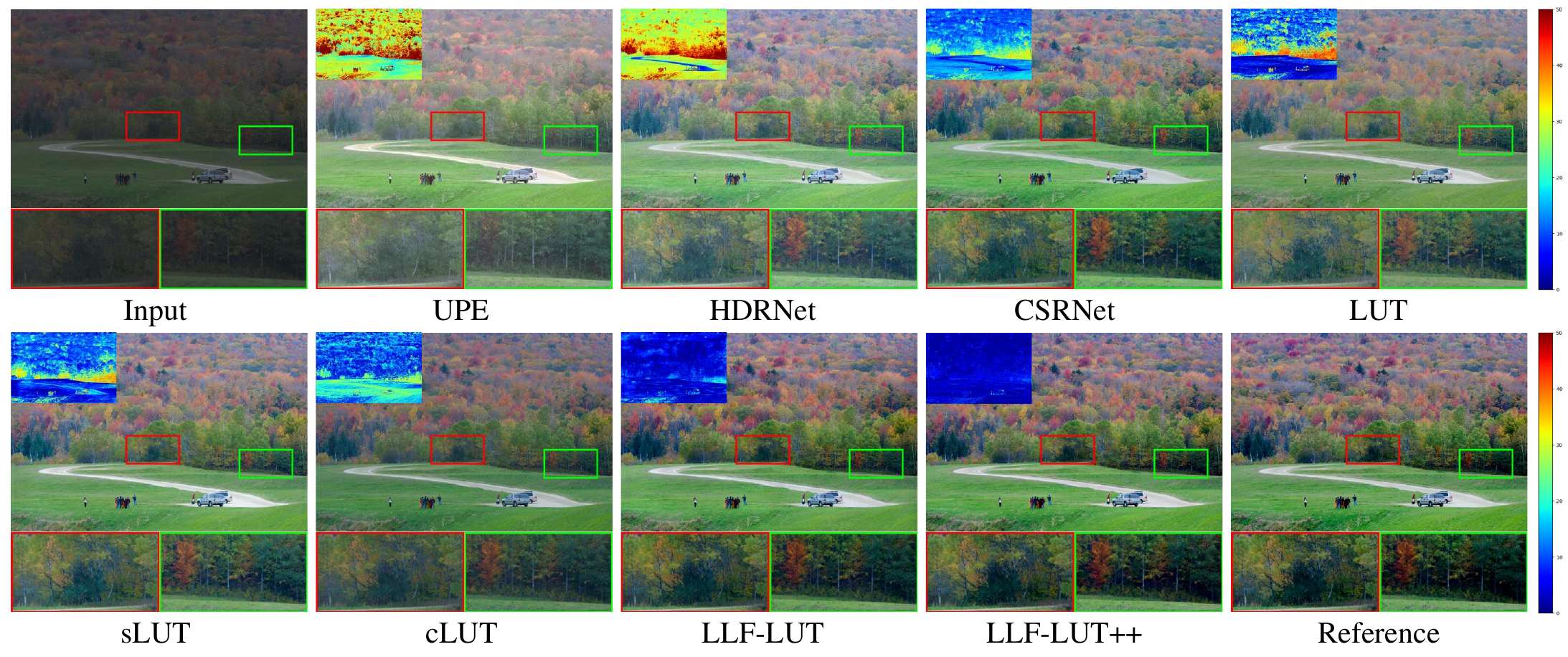}
    \vspace{-2em}
    \caption{Visual comparison with state-of-the-art methods on an original resolution test image from the MIT-Adobe FiveK dataset~\cite{bychkovsky2011learning}. The error maps in the upper left corner facilitate a more precise determination of performance differences. Best viewed in color and by zooming in.}    
    \label{fig:results_4K_fivek}
\end{figure*}

\begin{figure*}
    \centering
    \includegraphics[width=1\linewidth]{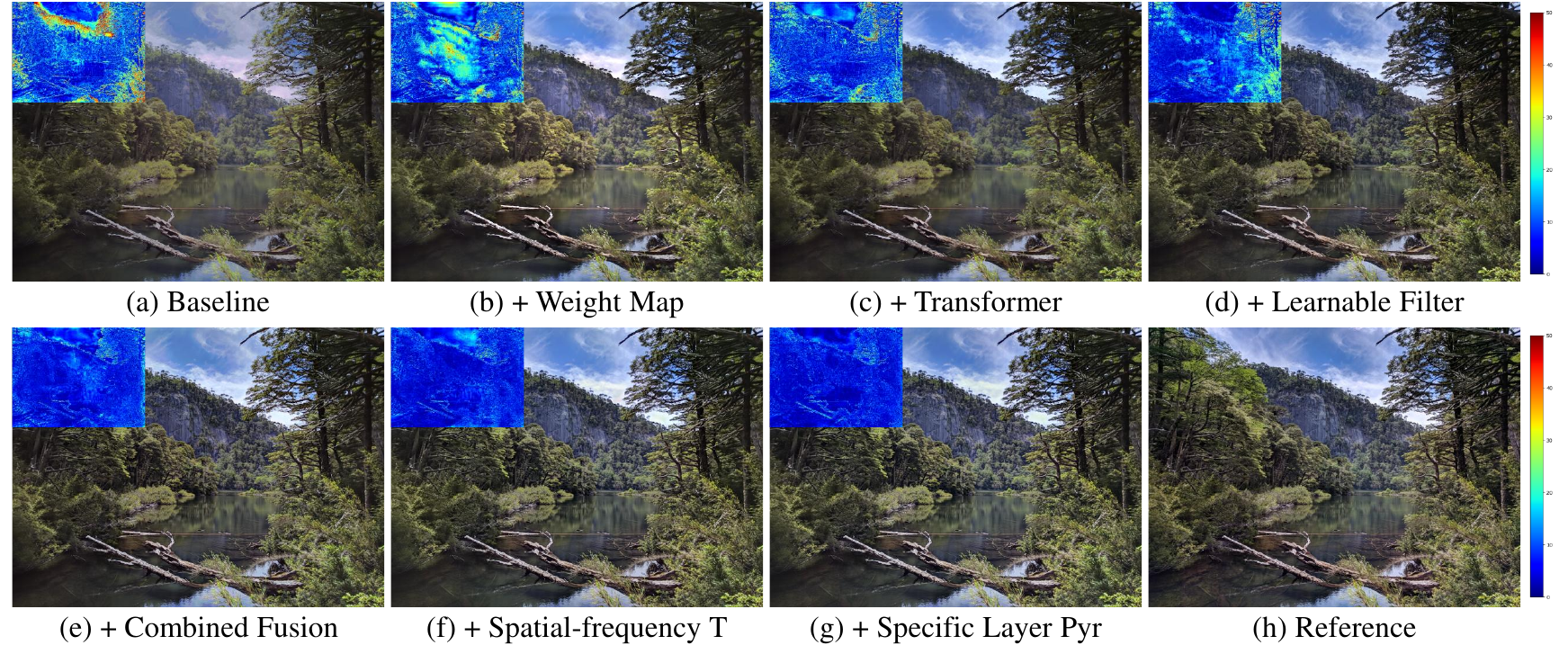}
    \vspace{-2em}
    \caption{
    Visual results of ablation study on framework component. (a) is the baseline method 3D LUT. (b) applies the pixel-level weight map. (c) implements the transformer backbone for weight prediction. (d) utilizes the learnable local Laplacian filter. (e) uses the combined weight fusion strategy, and (f) is with the proposed spatial-frequency transformer network. (g) deploys the specific layer pyramid network for reconstruction. And (h) is the reference. The error maps in the upper left corner facilitate a more precise determination of performance differences.
    }
    \label{fig:break_down}
\end{figure*}

\vspace{-0.5em}
\subsection{Quantitative Comparison Results}

Tab.~\ref{table:hdr} presents the quantitative comparison results on the HDR+ dataset for two different resolutions. Notably, our LLF-LUT++ exhibits a significant performance advantage over all competing methods on both resolutions, as indicated by the values highlighted in bold across all metrics. Specifically, LLF-LUT++ achieves a notable 0.58 dB improvement in PSNR compared to the best of other methods, DPRNet~\cite{yang2024learning}, at 480p resolution.
Besides, compared with DPRNet, our LLF-LUT++ demonstrates comparable performance at the original image resolution while requiring less runtime, demonstrating the robustness of our approach for high-resolution images.
Furthermore, even though the novel transformer contains few parameters, LLF-LUT++ surpasses LLF-LUT 1.81 dB and 2.64 dB at 480p and original resolution, respectively. This demonstrates that exploring the spatial-frequency information jointly can produce better weights for 3D LUTs fusion and benefit the image enhancement task. 

When evaluated on the MIT-Adobe FiveK dataset (refer to Tab.~\ref{table:mit}), our methods consistently demonstrate a clear advantage over all competing methods.
One can see that for all methods, LLF-LUT++ outperforms all competing methods on 480p and original resolution by at least 0.47 dB and 0.27 dB in terms of PSNR, respectively. Notably, compared to LLF-LUT, the proposed transformer in LLF-LUT++ leads to an improvement by 0.53 dB and 0.32 dB in terms of PSNR with fewer parameters. This further validates the effectiveness of our proposed LLF-LUT++.

\subsection{Qualitative Comparison Results}

To evaluate our proposed network intuitively, we visually compare enhanced images on the two benchmarks. Note that the input images are 16-bit TIF images, which regular display devices can not directly visualize; thus, we compress the 16-bit images into 8-bit images for visualization.

Fig.~\ref{fig:results_480p_hdr} and Fig.~\ref{fig:results_4K_hdr} present the result comparisons on the HDR+ dataset for two different resolutions.
In Fig.~\ref{fig:results_480p_hdr}, our method excels in preserving intricate details such as grass texture while enhancing brightness. Additionally, in the red box, LLF-LUT++ produces sharp edges whereas some halo artifacts are included in the results of other methods.
On the original resolution, the result of LLF-LUT++ exhibits superior color fidelity and alignment with the reference image. Specifically, compared to other methods, LLF-LUT++ can well enhance the challenge scenario in the red box with the lamp reflection cast upon the water surface.

The results on the MIT-Adobe FiveK dataset for two different resolutions are shown in Fig.~\ref{fig:results_480p_fivek} and Fig.~\ref{fig:results_4K_fivek}.
While other methods suffer from poor saturation in the hazed area of the sky, as shown in Fig.~\ref{fig:results_480p_fivek}, our method accurately reproduces the correct colors, resulting in a natural outcome.
Moreover, though the rich details of Fig.~\ref{fig:results_4K_fivek} complicate the task, LLF-LUT++ achieves a visually pleasing outcome with sharp edges and vivid colors.

Overall, our proposed network consistently delivers visually appealing results on both MIT-Adobe FiveK and HDR+ datasets. These findings highlight the effectiveness and superiority of our method in the enhancement task.

\subsection{Runtime Analysis}
To better characterize our model's computational complexity and efficiency, we provide the MACs and runtime of methods. Runtime values are averaged over 1000 images, each with dimensions of $3840\times2160\times3$, using a 32GB NVIDIA V100 GPU. The results are shown in Table.~\ref{table:hdr} and Table.~\ref{table:mit}.
The results indicate that the extensive computational demands of CSRNet and DeepLPF lead to slow processing. These methods, being pixel-level, are heavily reliant on hardware capabilities, particularly when handling large-resolution images. Conversely, our LLF-LUT++ balances computational complexity and performance.

Additionally, compared to LLF-LUT, the proposed LLF-LUT++ incorporates a transformer and a Laplacian pyramid network, leading to improved performance. Despite an increase in MACs, LLF-LUT++ improves PSNR by 2.64 dB and 0.32 dB on two high-resolution benchmarks while reducing runtime by 7.01 ms. Moreover, though other methods may have shorter runtimes, LLF-LUT++ reaches a significant performance enhancement of at least 3.93 dB and 0.57 dB in terms of PSNR on the two high-resolution benchmarks, sacrificing only a modest amount of computational resources. Overall, our approach achieves real-time high-quality enhancement of 4K high-resolution photographs.

\begin{figure*}
    \centering
    \includegraphics[width=1\linewidth]{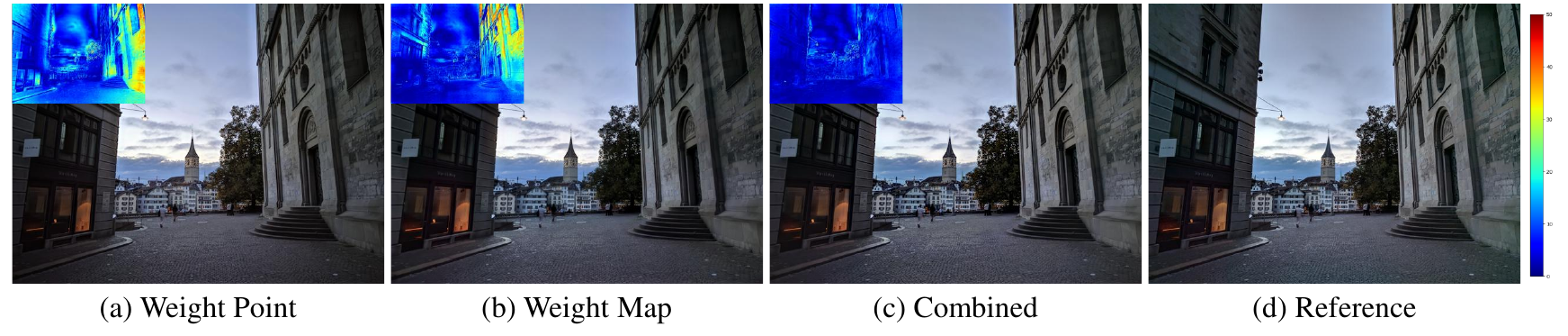}

    \vspace{-1em}
    \caption{Visual results of ablation study on weight fusion strategy. (a) and (b) use the weight point and weight map, respectively. (c) combines both weight point and weight map. (d) is the reference. The error maps in the upper left corner facilitate a more precise determination of performance differences.}
    \vspace{-0.5em}
    
    \label{fig:weight_strategy}
\end{figure*}

\subsection{Ablation Studies}
We perform ablation studies to demonstrate the effectiveness of each component of LLF-LUT++. Moreover, we study the effect of the basis 3D-LUT fusion strategy. Additionally, the study of the selection of pyramid layers is carried out. All experiments are conducted at the 480p resolution on the HDR+ dataset~\cite{hasinoff2016burst}.

\begin{table}
    \caption{Ablation study of framework component.}
    \vspace{-1em}
    \setlength{\tabcolsep}{8pt}
    \centering
        \begin{tabular}{l c c  c c }
            \toprule[0.15em]
            \multirow{2}{*}{\makecell[c]{Framework \\ Component}} &\multicolumn{4}{c}{Metrics}\\
            \cmidrule(r){2-5}
            & PSNR$\uparrow$ & SSIM$\uparrow$  & LPIPS$\downarrow$  & $\triangle E$$\downarrow$ \\
            \midrule[0.15em]
            Baseline           &23.16 &0.842 &0.113 &7.04 \\
            + Weight Map       &24.41 &0.856 &0.111 &6.23 \\
            + Transformer      &25.34 &0.868 &0.101 &5.92 \\
            + Learnable filter &26.62 &0.907 &0.063 &5.31 \\
            + Combined Fusion &27.09 &0.908 &0.071 &5.33   \\
            + Spatial-frequency T &27.94 &0.919 &0.057 &4.74 \\
            + Specific Layer Pyr &\textbf{28.43} &\textbf{0.924} & \textbf{0.056} &\textbf{4.54} \\
            \bottomrule[0.15em]
    \end{tabular}
    \label{table:ablation_break}
\end{table}

\vspace{0.5em}
\noindent
\textbf{Break-down ablations.} We conduct comprehensive breakdown ablations to evaluate the effects of our proposed approach.
The quantitative results are presented in Tab.~\ref{table:ablation_break}.

We begin with the baseline method, 3D LUT~\cite{zeng2020learning}, without utilizing pixel-level weight maps or learnable local Laplacian filters. The results show a significant degradation, indicating the insufficiency of 3D LUT.  When pixel-level weight maps are introduced, the results improve by an average of 1.25 dB. This evidence highlights the successful implementation of the pixel-level basis 3D LUTs fusion strategy discussed in Sec.~\ref{sec:fusion}.
Next, we replace the regular lightweight CNN backbone with a tiny transformer backbone proposed by~\cite{li2023efficient}, which contains less than 400K parameters. After deploying the transformer backbone, the model is improved by 0.93 dB, suggesting that the transformer backbone is more in line with global tone manipulation and benefits generating more visual pleasure LDR images.
Furthermore, when employing the image-adaptive learnable local Laplacian filter, the results exhibit a significant improvement of 1.28 dB. This finding indicates that the image-adaptive learnable local Laplacian filter facilitates the production of more vibrant results.
Moreover, employing the combined weight fusion strategy leads to a 0.47 dB improvement, which is further studied in detail in the following ablation study experiment.
Additionally, when replacing the tiny transformer backbone with our proposed spatial-frequency transformer, a significant advancement of 0.85 dB demonstrates the effectiveness of exploring both spatial and frequency information for this task.
Besides, adopting the specific layer pyramid network for reconstruction results in a 0.49 dB improvement.

As shown in Fig.~\ref{fig:break_down}, combining local Laplacian filter with 3D LUT achieves good visual quality on both global and local enhancement in this challenging case.
These results convincingly demonstrate the superiority of our proposed framework in photo enhancement tasks.

\vspace{0.5em}
\noindent
\textbf{Effect of basis 3D-LUT fusion strategy.}
To study the effect of basis 3D-LUT fusion strategy, we conduct experiments with weight point, weight map and combined weight, respectively.
Specifically, when using weight point, both HR and LR inputs are enhanced with the same weight point. In the case of weight map, the learned weight maps are applied to the LR inputs and up-sampled for enhancing HR inputs.

Tab.~\ref{table:ablation_fusion_strategy} presents the comparison results. There are two findings.
First, utilizing weight map yields better results than weight point by 2.23 dB PSNR. This confirms the superiority of weight map fusion strategy in accurately representing such pixel transformations.
Second, combining both weight point and weight map improves PSNR results by 0.92 dB, and achieves a balance between computational complexity and performance. In contrast to employing the same weight map, using different weights leads to specialized enhancements for HR and LR inputs. Utilizing weight map for LR inputs leads to high-quality results, providing a good start for pyramid reconstruction. Meanwhile, adopting a separate weight point for HR inputs can achieve efficient and stable pre-enhancement. With the pyramid network, the details of HR images can be well enhanced.
Additionally, Fig.~\ref{fig:weight_strategy} shows that the combined strategy leads to more natural output and enhances more details in the dark area. This further demonstrates the effectiveness of the combined fusion strategy.

\begin{table}
    \caption{Ablation study of basis 3D-LUT fusion strategy.}
    \vspace{-1em}
    
    \setlength{\tabcolsep}{10pt}
    \centering
        \begin{tabular}{c c c  c c }
            \toprule[0.15em]
            \multirow{2}{*}{Weight Strategy} &\multicolumn{4}{c}{Metrics}\\
            \cmidrule(r){2-5}
            & PSNR$\uparrow$ & SSIM$\uparrow$  & LPIPS$\downarrow$  & $\triangle E$$\downarrow$ \\
            \midrule[0.15em]
            Weight Point &25.28 &0.906 &0.068 & 5.93 \\
            Weight Map &27.51 & 0.912 & 0.066 & 4.99 \\
            Combined &\textbf{28.43} &\textbf{0.924} & \textbf{0.056} &\textbf{4.54} \\
            \bottomrule[0.15em]
    \end{tabular}
    \vspace{-0.5em}
    
    \label{table:ablation_fusion_strategy}
\end{table}

\begin{table}
    \caption{Ablation study of the selection of pyramid layers.}
    \vspace{-1em}
    
        \setlength{\tabcolsep}{9pt}
    \centering
        \begin{tabular}{c c c c c}
            \toprule[0.15em]
            \multirow{2}{*}{\makecell[c]{Low-frequency \\ Image Resolution}} &\multicolumn{4}{c}{Metrics}\\
            \cmidrule(r){2-5}
            & PSNR$\uparrow$ & SSIM$\uparrow$ &LPIPS$\downarrow$ &$\triangle E$$\downarrow$ \\
            \midrule[0.15em]
            $16\times16$ &27.12 &0.903 &0.078 &4.95\\
            $32\times32$ &27.68 &0.915 &0.063 &4.81\\
            $64\times64$ &28.43 &0.924 &0.056 &4.54\\
            $128\times128$ &28.49 &0.926 &0.058 &4.52\\
            $256\times256$ &28.51 &0.927 &0.059 &4.51 \\
            \bottomrule[0.15em]
    \end{tabular}
    \vspace{-1em}
    
    \label{table:ablation_pyrnum}
\end{table}

\vspace{0.5em}
\noindent
\textbf{Selection of the pyramid layers.}
We validate the influence of the number of Laplacian pyramid layers in LLF-LUT++. Our approach employs an adaptive Laplacian pyramid, allowing us to manipulate the number of layers by altering the resolution of the low-frequency image $I_{low}$. As shown in Tab.~\ref{table:ablation_pyrnum}, the model performs best on all evaluation metrics when the resolution of $I_{low}$ is set to $256\times256$. 

However, the proposed framework requires more computation. A trade-off between computational load and performance is determined by the number of layers in the Laplacian pyramid. The proposed framework remains robust when the resolution is reduced to alleviate the computational burden. For example, reducing the resolution of $I_{low}$ from $256\times256$ to $64\times64$ only marginally decreases the PSNR of the proposed framework from 28.51 dB to 28.43 dB. Remarkably, this reduction in resolution leads to a significant $30\%$ decrease in computational burden. These results validate that the tone attributes are presented in a relatively low resolution.

{Moreover, while a fixed input size is adopted during training, variable dimensions may occur during testing. In cases where the image downsampled in the bottom pyramid layer does not reach the required size of $64\times64$, we apply a specific upsampling operation to adjust the resolution of this layer to $64\times64$, while the dimensions of all other pyramid levels remain unchanged.}

\section{Limitation}
Photo enhancement is a fundamental task in the camera imaging pipeline. It has been studied for decades and is widely applied to cinematic color grading, computer graphics, and image processing. High dynamic range (HDR) imaging technology is gradually becoming ubiquitous as advancements are made in shooting equipment, rendering methods, and display devices. Hence, photo enhancement techniques tackling this issue are worth investigating. The proposed framework addresses the lack of local edge detail prevalent in learning-based photo enhancement methods. 

The learning-based photo enhancement methods mimic the retouching style of human photographers based on learned statistics of the training datasets. Consequently, these approaches may reflect any biases in the datasets, including those with negative social implications. The proposed framework presents this negative foreseeable societal consequence, either. Moreover, the subjective nature of photo enhancement tasks means color transformations may not universally satisfy all users. A plausible mitigation strategy is to customize fine-tuning to various user preferences.

Although our model demonstrates commendable efficacy concerning memory utilization and enhancement performance, it exhibits higher MACs and slower running time relative to the baseline method 3D LUT due to the constraints of the local Laplacian filter. However, this is a common limitation in local photo enhancement operators. Fundamentally, this is because local operators are inherently more complex than global operators, and their effects vary on a pixel-by-pixel basis, depending on the local image characteristics. Nevertheless, since human visual perception is primarily sensitive to local contrast, local operators can provide superior performance. Due to its high degree of data parallelism, the local Laplacian filter \cite{paris2011local,aubry2014fast} can efficiently leverage multi-core architectures. Using OpenMP, we have achieved substantial GPU acceleration, thereby reducing runtime to the point where 4K image processing can be accommodated. This represents a significant advancement in addressing the running speed as mentioned above constraints and improving the overall performance of the model.

{Directly applying our method to video on a frame-by-frame basis may lead to temporal inconsistencies, such as flickering artifacts, due to the lack of explicit temporal modeling. This is a common challenge for image-based enhancement methods, and our approach is no exception. While our current work focuses on image enhancement, the pyramid architecture and locally adaptive LLF-LUT++ design could potentially be extended to video by incorporating temporal constraints.  In the future work, we plan to explore temporal consistency mechanisms such as optical-flow-guided feature alignment and temporal smoothing techniques to mitigate flickering artifacts while maintaining computational efficiency.}

\section{Conclusion}
This paper proposes an effective end-to-end framework LLF-LUT++ for photo enhancement tasks, combining global and local enhancements. The proposed framework utilizes the Laplacian pyramid decomposition technique to handle high-resolution images effectively. This approach significantly reduces computational complexity while simultaneously ensuring uncompromised enhancement performance. It combines the advantages of weight map fusion and weight point fusion, enhancing both high-resolution (HR) and low-resolution (LR) inputs. Moreover, it introduces a novel weight prediction transformer network that fuses spatial and frequency information. An image-adaptive learnable local Laplacian filter is proposed to progressively refine the high-frequency components, preserving local edge details and reconstructing the pyramids. The proposed method achieves real-time processing of 4K images, with a processing time of only 13 ms per image. Extensive experimental results on two publically available benchmark datasets show that our model performs favorably against state-of-the-art methods for both 480p and 4K resolutions.

\bibliography{reference}
\bibliographystyle{ieeetr}
\vspace{-4em}

\begin{IEEEbiography}
[{\includegraphics[width=1in,height=1.25in,clip,keepaspectratio]{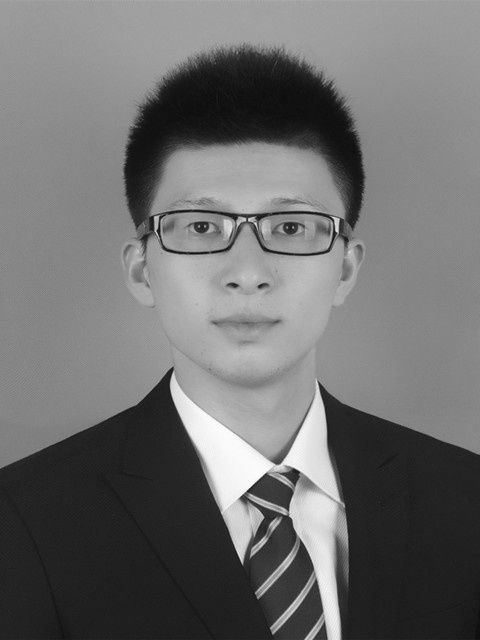}}]{Feng Zhang}
is currently a Ph.D. candidate in the School of Artificial Intelligence and Automation, Huazhong University of Science and Technology, supervised by Prof. Nong Sang. He received his M.S. degree from the School of Materials Science and Engineering, Huazhong University of Science and Technology in 2015. His research interests include computer vision and deep learning. Now he mainly works on the area of low-light image enhancement.
\end{IEEEbiography}
\vspace{-2em}

\begin{IEEEbiography}
[{\includegraphics[width=1in,height=1.25in,clip,keepaspectratio]{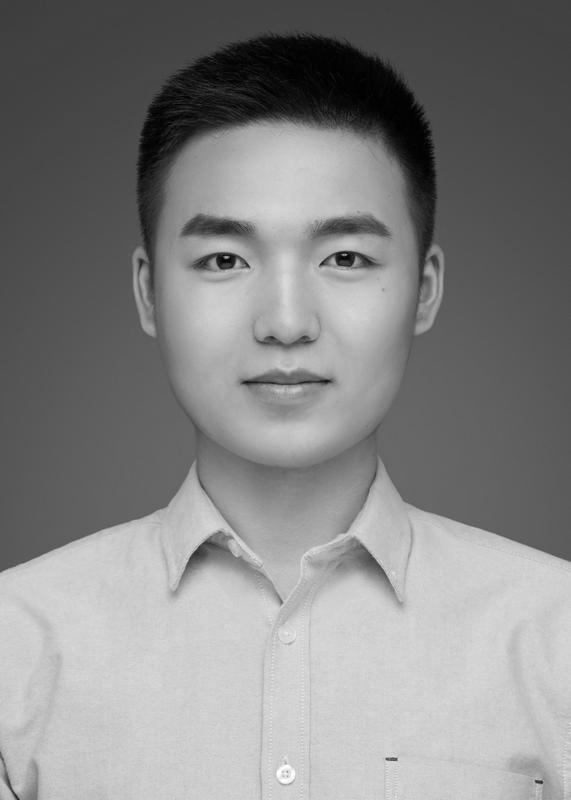}}]{Haoyou Deng}
received the B.S. degree in Huazhong University of Science and Technology, China, in 2021. He is currently pursuing the Ph.D. degree in artificial intelligence from Huazhong University of Science and Technology, China. His research interests include image processing and image/video restoration.
\end{IEEEbiography}
\vspace{-2em}

\begin{IEEEbiography}
[{\includegraphics[width=1in,height=1.25in,clip,keepaspectratio]{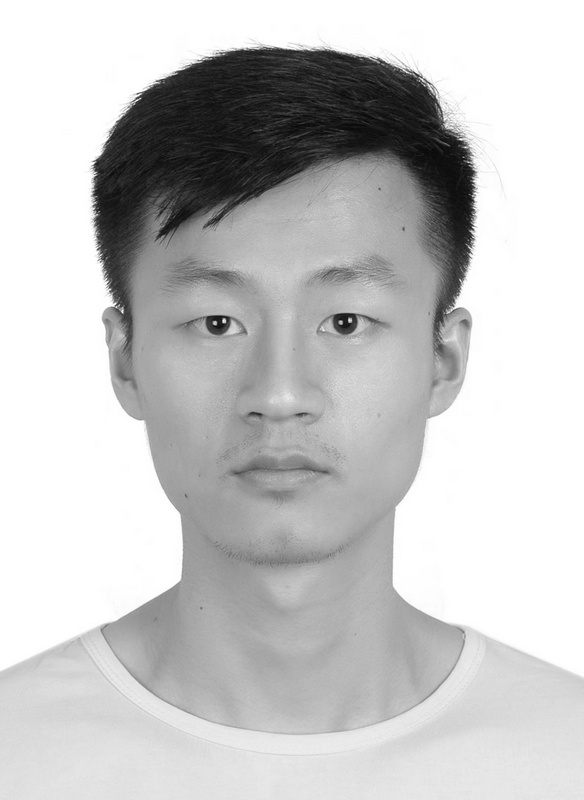}}]{Zhiqiang Li}
is currently pursuing a Ph.D. at Huazhong University of Science and Technology. He graduated with a bachelor’s degree from Xidian University in 2016 and obtained his master’s degree from Huazhong University of Science and Technology. His research interests include image signal processing and machine learning.
\end{IEEEbiography}
\vspace{-2em}

\begin{IEEEbiography}
[{\includegraphics[width=1in,height=1.25in,clip,keepaspectratio]{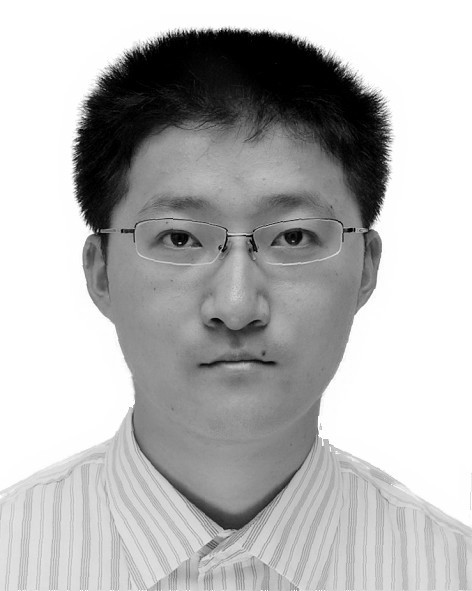}}]{Lida Li}
received the Ph.D. degree from the Department of Computing, The Hong Kong Polytechnic University in 2021 and the B.S. and M.Sc. degrees from the School of Software Engineering, Tongji University, Shanghai, China, in 2013 and 2016, respectively. He is currently with DJI. He has published papers in top journals and top conferences. His research interests are machine learning and image processing.
\end{IEEEbiography}
\vspace{-2em}

\begin{IEEEbiography}
[{\includegraphics[width=1in,height=1.25in,clip,keepaspectratio]{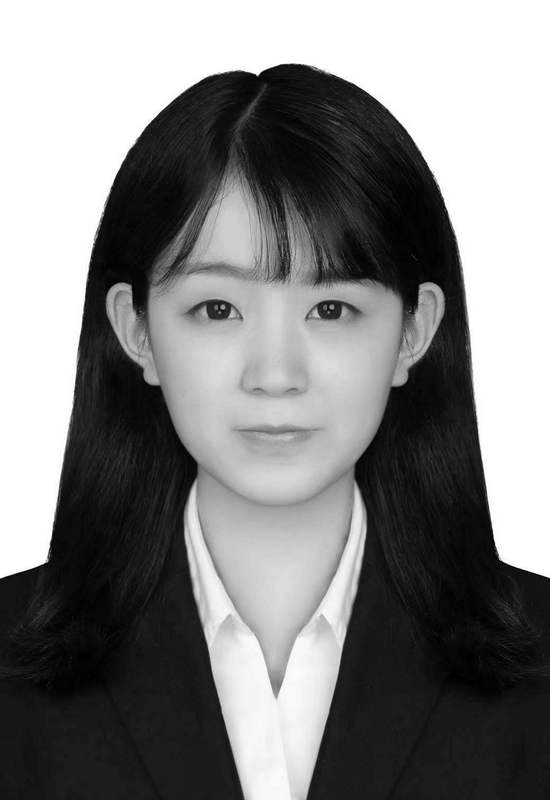}}]{Bin Xu}
is a member of imaging group of DJI currently. She received her B.S. and M.S. degrees from Nanjing University Of Aeronautics and Astronautics in 2015 and 2018 respectively. Her current research interests include image processing and machine learning.
\end{IEEEbiography}
\vspace{-4em}

\begin{IEEEbiography}
[{\includegraphics[width=1in,height=1.25in,clip,keepaspectratio]{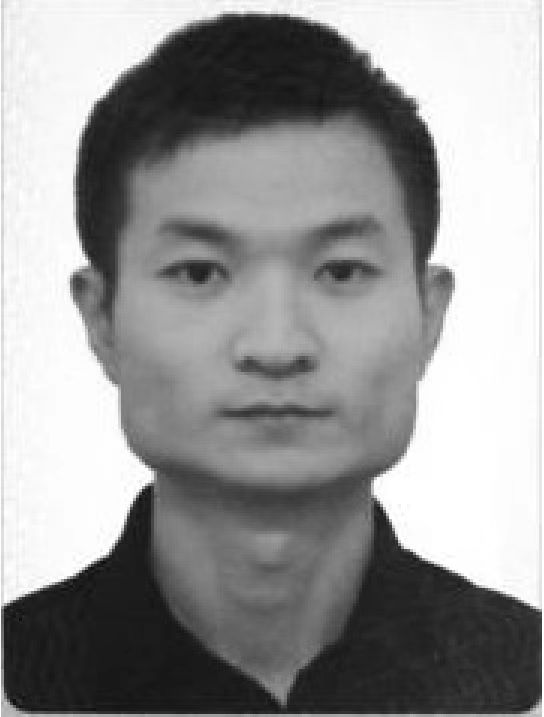}}]{Qingbo Lu}
received the B.S. degree from the School of Information Science and Engineering from the Southeast University, Nanjing, China, in 2011 and the Ph.D. degree in electronic engineering and information science from the University of Science and Technology of China, in 2016. He is currently with DJI. His current research interests include image/video processing, sparse representation, image quality assessment, and computer vision.
\end{IEEEbiography}
\vspace{-2em}

\begin{IEEEbiography}[{\includegraphics[width=1in,height=1.25in,keepaspectratio]{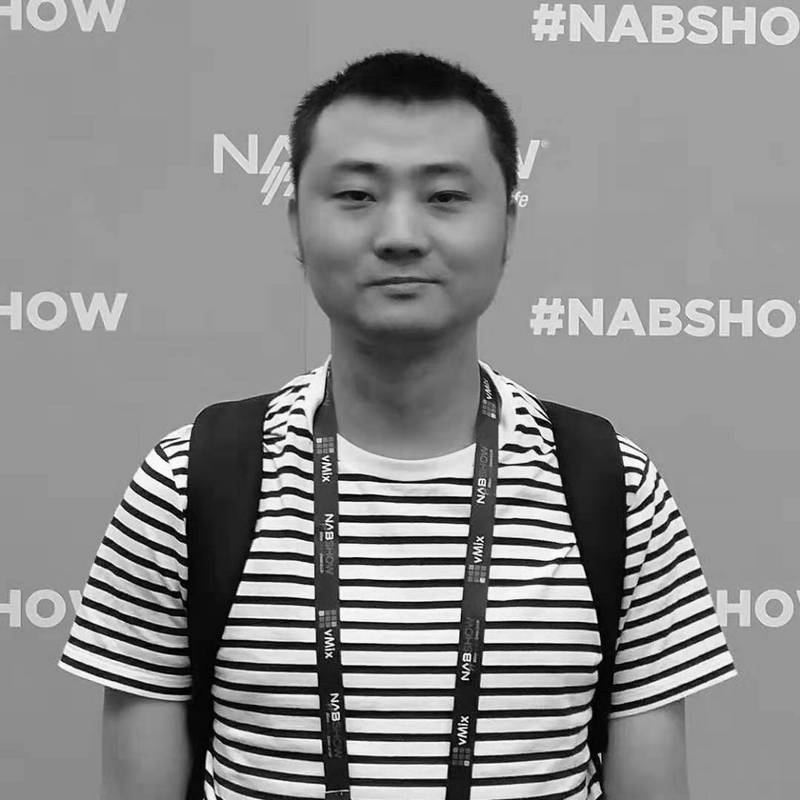}}]{Zisheng Cao} is currently with imaging group of DJI. He received his B.S. and M.S. degrees from Tsinghua University in 2005 and 2007, respectively, and Ph.D. degree from the University of Hong Kong in 2014. His research interests include image signal processing and machine learning. Before joining DJI, he was a research scientist in Philips.
\end{IEEEbiography}
\vspace{-4em}

\begin{IEEEbiography}
[{\includegraphics[width=1in,height=1.25in,clip,keepaspectratio]{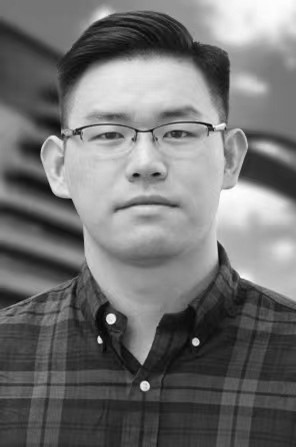}}]{Minchen Wei} is a professor at The Hong Kong Polytechnic University. He obtained his bachelor degree at Fudan University, Master's and Ph.D. degrees at The Pennsylvania State University. His research mainly focuses on color science, imaging science, and their application on multi-media systems.
\end{IEEEbiography}
\vspace{-2em}

\begin{IEEEbiography}
[{\includegraphics[width=1in,height=1.25in,clip,keepaspectratio]{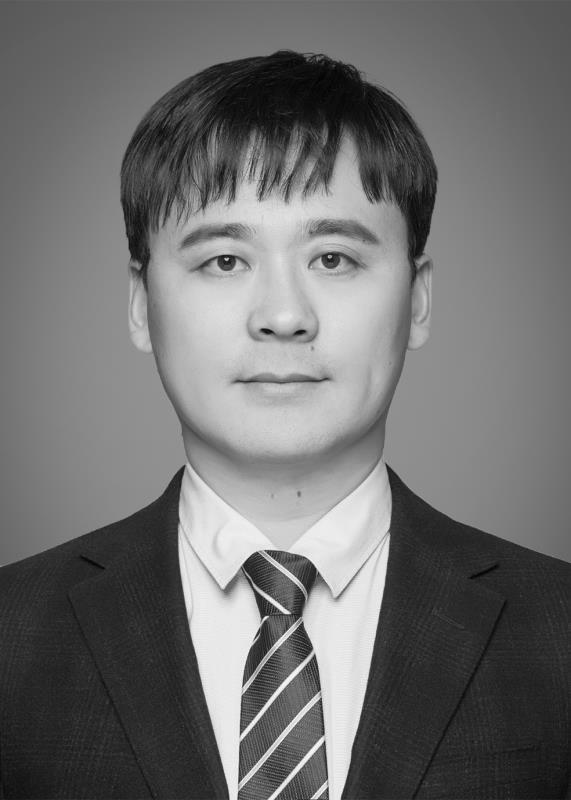}}]{Changxin Gao}
received the Ph.D. degree in pattern recognition and intelligent systems from the Huazhong University of Science and Technology in 2010. He is currently a Professor with the School of Artificial Intelligence and Automation, Huazhong University of Science and Technology. His research interests are pattern recognition and surveillance video analysis.
\end{IEEEbiography}
\vspace{-2em}

\begin{IEEEbiography}
[{\includegraphics[width=1in,height=1.25in,clip,keepaspectratio]{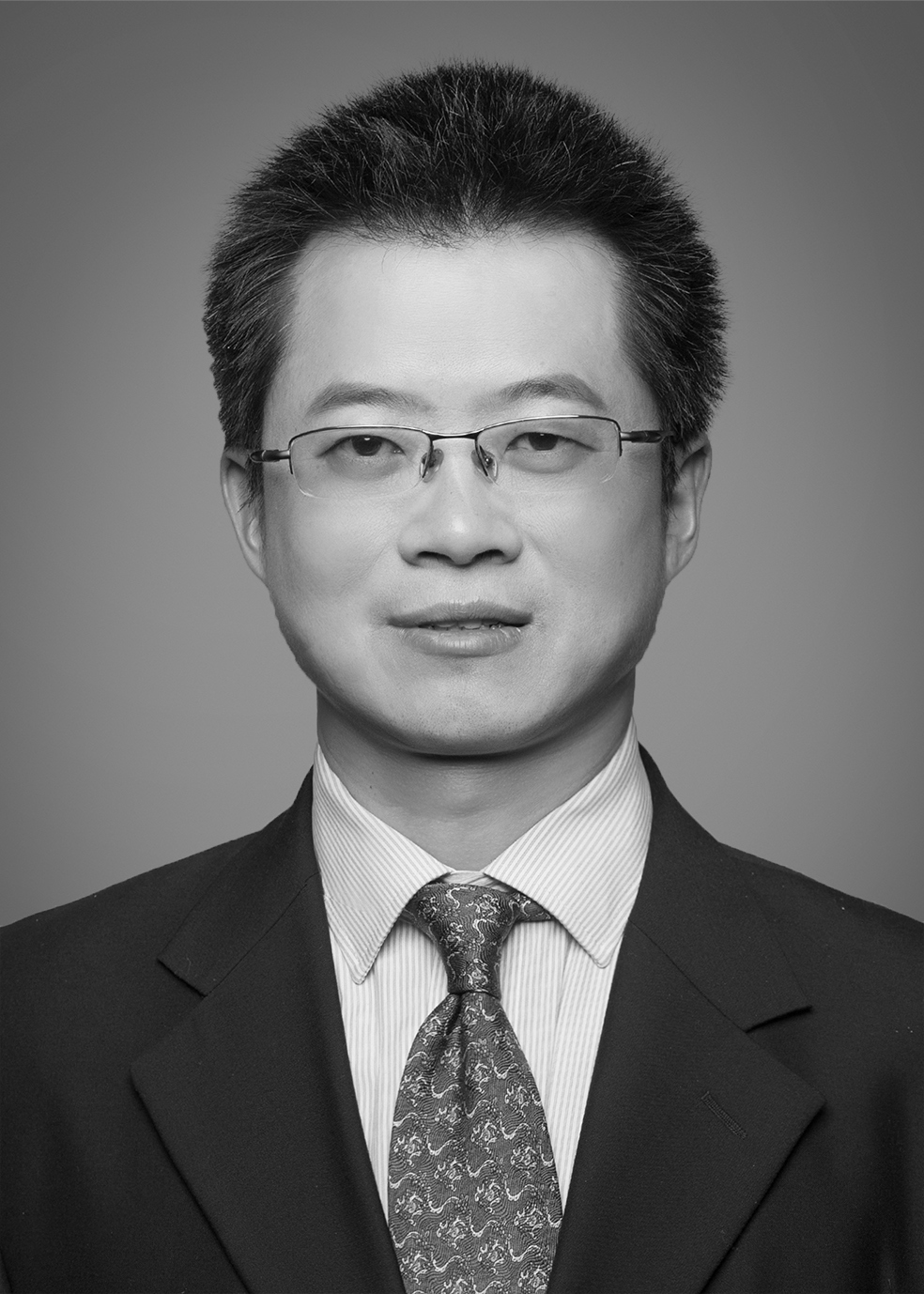}}]{Nong Sang}
received the Ph.D. degree in pattern recognition and intelligent control from the Huazhong University of Science and Technology in 2000. He is currently a Professor with the School of Artificial Intelligence and Automation, Huazhong University of Science and Technology. His research interests include pattern recognition, computer vision, and neural networks.
\end{IEEEbiography}
\vspace{-2em}

\begin{IEEEbiography}
[{\includegraphics[width=1in,height=1.25in,clip,keepaspectratio]{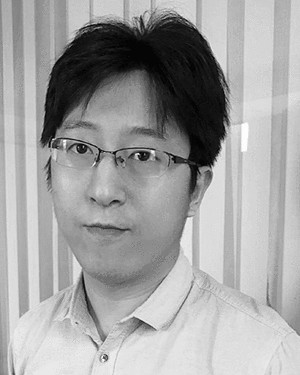}}]{Xiang Bai} (Senior Member, IEEE) received the B.S., M.S., and Ph.D. degrees in electronics and information engineering from the Huazhong University of Science and Technology (HUST), Wuhan, China, in 2003, 2005, and 2009, respectively. He is currently a professor with the School of Software Engineering, HUST. He is also the vice-director of the National Center of Anti-Counterfeiting Technology, HUST. His research interests include object recognition, shape analysis, and scene text recognition and intelligent systems. He serves as an associate editor with Pattern Analysis and Machine Intelligence, Pattern Recognition, Pattern Recognition Letters, Neurocomputing and Frontiers of Computer Science.
\end{IEEEbiography}
\vspace{-2em}

\balance

\end{document}